\documentclass[review]{elsarticle}
\usepackage{hyperref}
\usepackage{amssymb}
\usepackage{amsmath}
\usepackage{multirow}
\usepackage{url}
\usepackage{booktabs}
\usepackage{subfigure}
\usepackage{amssymb}
\usepackage{graphicx}
\usepackage{array}
\usepackage{paralist}
\usepackage{amssymb}
\usepackage{multirow}
\usepackage{ulem}
\usepackage{algorithmicx,algorithm}
\journal{Information Sciences}

\begin{document}

\begin{frontmatter}

\title{RTFN: A Robust Temporal Feature Network for Time Series Classification}

\author[address1]{Zhiwen Xiao}
\author[address2]{Xin Xu}
\author[address1]{Huanlai Xing\corref{mycorrespondingauthor}}

\cortext[mycorrespondingauthor]{Corresponding author}
\ead{hxx@home.swjtu.edu.cn}
\author[address1]{Shouxi Luo}
\author[address1]{Penglin Dai}
\author[address1]{Dawei Zhan}

\address[address1]{Southwest Jiaotong University, Chengdu, China}
\address[address2]{China University of Mining and Technology, Xuzhou, China}

\begin{abstract}

Time series data usually contains local and global patterns. Most of the existing feature networks pay more attention to local features rather than the relationships among them. The latter is, however, also essential yet more difficult to explore. To obtain sufficient representations by a feature network is still challenging. To this end, we propose a novel robust temporal feature network (RTFN) for feature extraction in time series classification, containing a temporal feature network (TFN) and an LSTM-based attention network (LSTMaN). TFN is a residual structure with multiple convolutional layers. It functions as a local-feature extraction network to mine sufficient local features from data. LSTMaN is composed of two identical layers, where attention and long short-term memory (LSTM) networks are hybridized. This network acts as a relation extraction network to discover the intrinsic relationships among the features extracted from different data positions. In experiments, we embed RTFN into a supervised structure as a feature extractor and an unsupervised structure as an encoder, respectively. The results show that the RTFN-based structures achieve excellent supervised and unsupervised performance on a large number of UCR2018 and UEA2018 datasets.

\end{abstract}


\end{frontmatter}

\section{Introduction}
Time series data has been used in various domains, such as weather forecasting \cite{F1}, traffic analysis \cite{F3,F2}, human activity recognition \cite{F4,F5}, clinical diagnosis \cite{F6}, human heart record \cite{F7}, electricity demand \cite{F8}, etc. Making full use of the data in real-world applications is crucial, depending on how well features are extracted. Different from other types of data, such as ImageNet for image classification \cite{F9}, SemEval-2014 for sentiment recognition \cite{F10}, and ICDAR2019 for natural scene text processing \cite{F11}, a time series is a sequence of time-ordered data points recording certain processes \cite{F12}. In time series data, local patterns are local temporal features, while global patterns are relationships among local ones. Recently, effective feature and relation extraction has become a critical challenge, which is also a basis for time series classification \cite{F12,F14,F13,F15}.

So far, there have been two classes of approaches for addressing the challenge above, including traditional algorithms and deep learning ones \cite{F14}. Traditional approaches aim at mining features and regularizations from data by revealing the significant differences and connections within the data. These approaches are mainly distance-based and feature-based. Distance-based approaches address a classification task by measuring the similarities between spatial features of data. Combining the nearest neighbor (NN) and the dynamic time warping (DTW) is widely adopted  \cite{F12,F16}, such as, $DD_{DTW}$ \cite{F71}, $DTD_C$ \cite{F71}, $DTW_I$ \cite{F18}, $DTW_{D}$ \cite{F53}, $DTW_{A}$ \cite{F15}, etc. Besides, there are a number of NN-DTW-based ensemble approaches. For example, the elastic ensemble (EE) uses 11 1-NN-based elastic distance measures to achieve decent performance on various time series datasets \cite{F16}. The collective of transformation ensembles (COTE) considers over 35 weighted classifiers \cite{F14}. In addition, the hierarchical vote collective of transformation-based ensembles (HIVE-COTE) \cite{F17} and the local cascade ensemble (LCE) \cite{F18} are two representative algorithms in the literature.

On the other hand, feature-based approaches pay more attention to exploring representative features from time series data. For instance, Baydogan et al. \cite{F74} introduce a bag-of-features framework for feature extraction via a truncated discrete Fourier transform. The hidden state conditional random field (HCRF) uses hidden variables to build the latent structure of the input \cite{F21}. Besides, the learned pattern similarity (LPS) \cite{F61}, the bag of SFA symbols (BOSS) \cite{F51}, the time series forest (TSF) \cite{F73}, and the hidden unit logistic model (HULM) \cite{F20} are also feature-based.

Apart from the distance- and feature-based approaches, research efforts have also been dedicated to other techniques. For example, the ultra fast shapelets (UFS) applies support vector machine and random forest to generate random shapelets from the input \cite{F59}. The symbolic representation for multivariate time series (SMTS) adopts random forest to divide time series data into leaf nodes for local-pattern extraction \cite{F76}. Tuncel et al. \cite{F60} apply autoregressive forests (mv-ARF) to mine representative shapelets from multivariate time series data. Karlsson et al. \cite{F55} use generalized random shapelet forests (gRSF) to extract significant features. WEASEL+MUSE utilizes a bag-of-patterns model with various sliding window sizes for multivariate feature extraction \cite{F56}.

On the other hand, the deep learning algorithms aim at unfolding the internal representation hierarchy of data, which helps to capture the intrinsic connections among representations \cite{F22}. These algorithms are roughly classified into two categories, namely single-network-based and dual-network-based models. To be specific, a single-network-based model uses one (usually hybridized) network to handle both feature and relation extraction. These models pay attention to mining the basic representation hierarchy of data and significant connections within the hierarchy, e.g., ConvTimeNet \cite{F23}, InceptionTime \cite{F24}, and OmniScale 1-dimensional convolutional neural network (OS-CNN) \cite{F25}. A dual-network-based model is composed of a local-feature extraction network and a relation extraction network in parallel. The first network, usually convolutional structure based, concentrates on local features. In contrast, the second one focuses on relationships among the features extracted, e.g. Transformer-based model \cite{F26}, ALSTM-FCN \cite{F27}, etc. Compared with feature extraction, relation extraction aims at capturing those hidden connections among the features extracted before. In other words, a relation extraction network is able to compensate for the loss of the representations ignored by its corresponding feature extraction network. Hence, it is  vital to design an effective relation extraction network for different applications \cite{F12,F14,F26}. Nowadays, attention and long short-term memory (LSTM) networks are widely used for relation extraction in time series classification. That is because the attention mechanism can relate different positions of a sequence to derive the relationships at certain positions \cite{F28} and LSTM is able to explore long- and short-period dependencies in data, both of which help to enhance the relation extraction \cite{F30,F29,F27}. 

Nowadays, hybridizing attention and LSTM networks for relation extraction has attracted increasingly more research attention in time series classification. There are mainly two ways to combine them, namely cascading and embedding models. A cascading model stacks attention and LSTM networks one after another to realize some specific functions. But, neither attention networks nor LSTM networks require significant changes in their structures, e.g. the attention LSTM (AttLSTM) models \cite{F27,F31}. However, the cascading models usually suffer from two drawbacks. Firstly, almost all attention networks are based on fully connected networks, which are not sensitive to the intricate features hidden in data. Secondly, the useful representations extracted before are easily lost as they go through subsequent networks. An embedding model, on the other hand, integrates attention and LSTM networks in a compact manner. With fewer layers of neural networks cascaded, fewer features are lost during data transmission. Thus, more useful features and relationships are made use of by the model. Such a model is sensitive to regular and periodic data and hence able to concentrate on the local and periodical variations of data, e.g., LSTM with trend attention gate (LSTMTAG) \cite{F32}. If not designed properly, an embedding model may not be aware of the global variations of those non-periodical data, especially when handling long univariate and multivariate datasets. Theoretically speaking, compared with fully connected networks, embedding LSTM networks into an attention structure helps to provide it with significantly more temporal features for calculation, improving its sensitivity to the global variations of non-periodical data. Unfortunately, such a structure has not been considered in the time series data mining community.

To take advantage of the dual-network-based model and the hybridization of attention and LSTM networks, we propose a robust temporal feature network (RTFN) for feature extraction in the area of time series classification. RTFN consists of a temporal feature network as its local-feature extraction network and an LSTM-based attention network as its relation extraction network. Our main contributions are summarized below.

\begin{itemize}
\item[-] The temporal feature network is a CNN-based residual structure, responsible for extracting sufficient local features. Multi-head CNN layers are used to diversify multi-scale features, and self-attention is adopted to relate different positions of the features extracted before. Besides, we use the leaky rectified linear unit as the activation function to reduce the loss of features during their transmission. 
\item[-] The LSTM-based attention network contains two identical layers. In each layer, instead of fully connected networks, LSTM networks are used to obtain the query, key, and value matrices for their corresponding attention structure. Unlike the existing structures that combine attention and LSTM networks, the LSTM-based attention network can pay attention to the global variations of non-periodical data, which helps to mine useful relationships among the features already learned.
\item[-] We embed RTFN into a supervised structure and test it on 85 UCR2018 datasets and 30 UEA2018 datasets, where the RTFN-based algorithm outperforms a number of existing supervised algorithms on 39 UCR2018 and 15 UEA2018 datasets, respectively. Also, we embed RTFN into a simple unsupervised clustering as an encoder. Our structure wins 9 out of 36 UCR2018 datasets, compared with 13 unsupervised algorithms.
\end{itemize}

The rest of the paper is organized as follows. Section 2 reviews the state-of-the-art deep learning algorithms for time series classification and various combinations of attention and LSTM networks. Section 3 introduces the overview of RTFN, its key components, the RTFN-based supervised structure, and the RTFN-based unsupervised clustering. Experimental results are given in Section 4. Section 5 concludes the paper.

\section {Related Work}
This section first reviews the deep learning algorithms for time series classification and then discusses the existing means to hybridize attention and LSTM networks. 
\subsection{Deep Learning Algorithms}
Since the introduction of the fully convolutional network (FCN) \cite{F33}, increasingly more algorithms have been proposed to address time series classification problems \cite{F14}. In general, these algorithms are either single-network-based or dual-network-based. Single-network-based algorithms focus on significant features of data. For example, a 34-layer convolutional neural network was constructed to handle the ECG classification problem \cite{F34}. Serrà et al. developed a universal encoder based on CNN and convolutional attention to mine the temporal representations from input data \cite{F35}. In \cite{F23}, an off-the-shelf deep CNN (ConvTimeNet) with four convolutional blocks was proposed as a transferable network to adapt quickly for the requirements of datasets. Fawaz et al. used a fast gradient sign method to fool a ResNet model, called adversarial attacks for time series classification, where a set of synthetic samples was generated \cite{F36}. Besides, InceptionTime \cite{F24}, and OS-CNN \cite{F25} are often regarded as two representative single-network-based models, achieving decent performance on many univariate time series datasets. InceptionTime uses an inception structure to explore multi-scale representations from data, and OS-CNN adopts 1-dimensional CNN to mine local features and the relationships among the data. On the other hand, as an emerging trend for time series classification, dual-network-based models have not yet received much research attention. A few LSTM-FCN-based models were designed to cope with univariate and multivariate time series classification problems\cite{F27,F31}, where FCN and LSTM were used for feature and relation extraction, respectively. In \cite{F26}, Huang et al. proposed a residual attention net (RAN) consisting of a ResNet-based feature network and a transformer-based relation network, obtaining promising performance on UCR datasets. Meanwhile, the dual-network-based algorithms usually achieved better classification performance than those single-network-based ones \cite{F26,F27,F31}.

\subsection{Hybridization of Attention and LSTM}
Hybridizing attention and LSTM models is an emerging solution to temporal and spatial relation extraction. The existing works mainly include cascading and embedding models. The former simply stacks attention and LSTM structures together. For instance, an attention-LSTM model was adopted to cope with univariate and multivariate time series classification on UCR and UEA datasets, respectively \cite{F27,F31}. An attention-LSTM model was integrated into the convolution–deconvolution word embedding to merge context-specific and task-specific information \cite{F37}. In \cite{F38}, an attention-based time-incremental CNN cascaded attention and LSTM networks for temporal and spatial information fusion of ECG signals. Besides, the cascading models have been applied to video segmentation \cite{F39}, semantic relation extraction \cite{F40}, visual question answer \cite{F41}, urban flow prediction \cite{F42}, and so on. On the other hand, the embedding models focus on the compact integration of attention and LSTM networks. For instance, a TAG-embedded LSTM model was devised to explore the local variations of quasi-periodic time series data \cite{F32}. Wang et al. \cite{F43} proposed a novel online attentional recurrent neural network (ARNN) model for video tracking, where inter- and intra-attention models were embedded into a bi-directional LSTM to distinguish different background scenarios. In \cite{F44}, Chen et al. put forward an identity-aware single shot multi-boxes detector for object detection, where an attention-embedded LSTM structure was used to locate positions of interesting objects. In \cite{F45}, an attention-based LSTM was introduced to capture the representation hierarchy of data in power consumption forecasting. 

\subsection{Analysis and Motivation}
The dual-network-based models realize feature and relation extraction by two separate networks in parallel. Such a model usually performs better than a single-network-based model in supervised classification and unsupervised clustering, according to references \cite{F26,F27,F31} and our observations in Section 4.4. Nevertheless, designing a dual-network-based model is quite challenging since its structure should meet datasets' requirements, especially its relation extraction network. The hybridization of attention and LSTM offers a promising means to discover the relationships among the representations obtained from data. But, the existing cascading and embedding models cannot well handle the global variations of non-periodical time series data. The above is the motivation why we design a dual-network-based algorithm for time series classification and why we embed LSTM into an attention structure for relation extraction.

\section{RTFN}
This section first overviews the structure of RTFN and then describes two important components, namely a convolutional neural block and an LSTM-based attention layer. In the end, the RTFN-based supervised structure and unsupervised clustering are introduced, respectively.

\begin{figure}[htbp]
  \centering
 \label{miain}
\includegraphics[width=12cm]{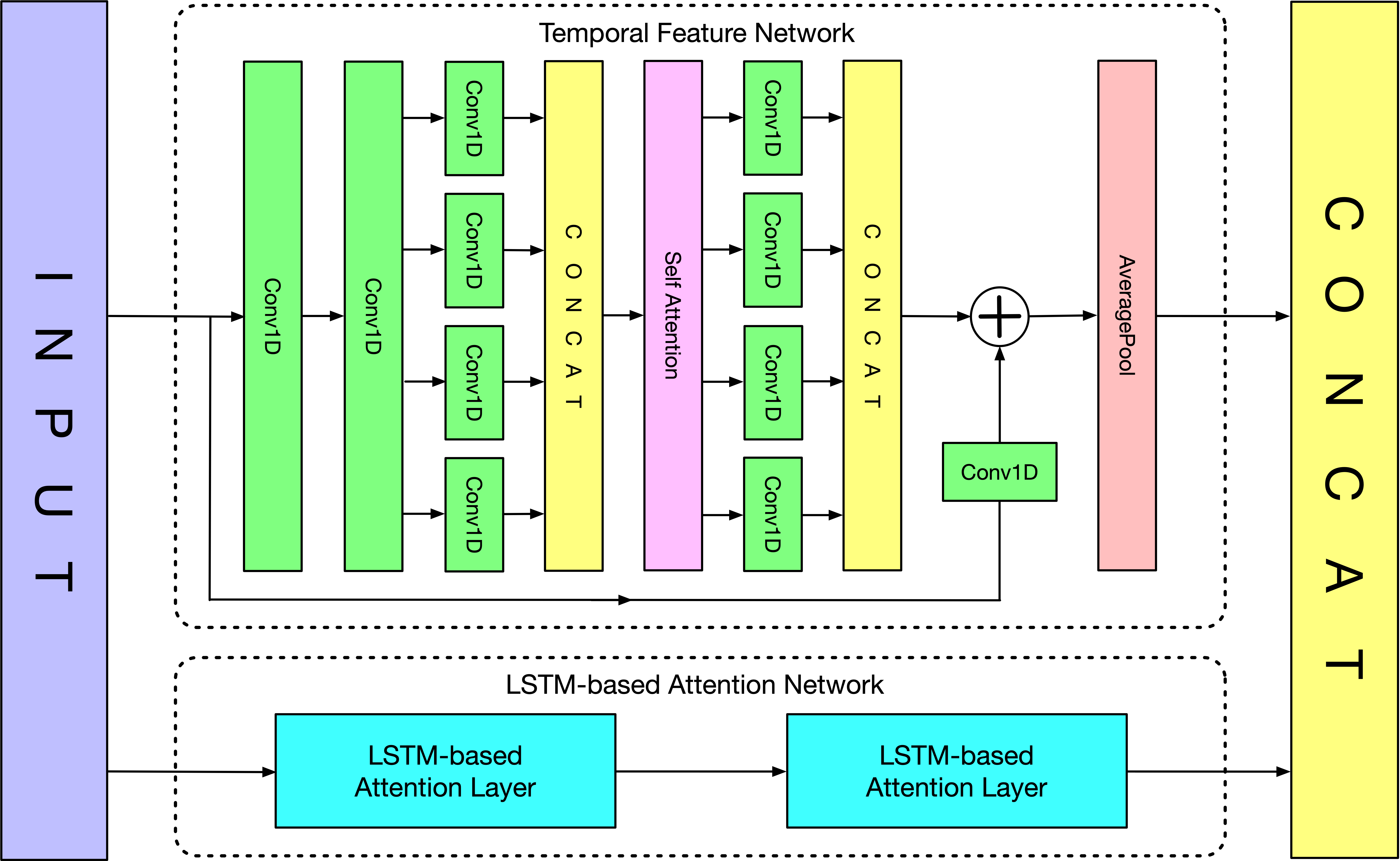}
\caption{Structure of RTFN.}
\end{figure}

\subsection{Overview}

The structure of RTFN is shown in Fig. 1. It primarily consists of a temporal feature network (TFN) and an LSTM-based attention network (LSTMaN). In TFN, a convolutional neural block, namely 'Conv1D', is seen as a basic building block responsible for capturing the local features from the input. Two multi-head convolutional neural layers, each consisting of four Conv1D blocks, are used to discover higher-level multi-scale features from the lower-level features extracted before. Besides, we place a self-attention layer \cite{F28} between the two multi-head layers to relate the positions of the local features obtained by the first multi-head layer, which enriches the input features of the second multi-head layer. Detailed observations can be found in Section 4.3. LSTMaN is composed of two LSTM-based attention layers, aiming at digging out the intrinsic relationships among the features learned from the input, which helps to compensate for the loss of the representations ignored by TFN. TFN and LSTMaN are used as the local-feature and relation extraction networks, respectively. Combining them provides RTFN with sufficient features and relationships. In this paper, RTFN is embedded into a supervised structure in Section 3.4 and an unsupervised clustering in Section 3.5, respectively.

\subsection{Convolutional Neural Block (Conv1D)}
A Conv1D block consists of a 1-dimensional CNN module, a batch normalization module, and a leaky rectified linear unit (LeakReLU) activation function \cite{F46}, as defined in Eq. (1).
$$O_{Conv1D}=f_{LeakReLU}(f_{BN}(f_{conv}(x)))\eqno{(1)}$$
where, $O_{Conv1D}$ and $x$ are the output and input of the Conv1D block, respectively. $f_{LeakReLU}$, $f_{BN}$, and $f_{conv}$ denote the $LeakReLU$ activation, batch normalization, and CNN functions, respectively. 

The CNN module is used to explore the local features from the input \cite{F22}, as defined in Eq. (2).
$$f_{conv}(x)= W_{cnn}\otimes x + b_{cnn}\eqno{(2)}$$
where, $W_{cnn}$ and $b_{cnn}$ are the weight and bias matrices of CNN, respectively. $\otimes$ is the convolutional computation operation. 

Let $x_{bn}=\{a_1,a_2,...,a_{N}\}$ be the input of the batch normalization module, where $a_i$ and $N$ are the $i$-th instance and the batch size, respectively. Let $\mu=\frac{1}{N}\sum_{i=1}^{N}{a_i}$ and $\delta=\sqrt{\frac{1}{N}\sum_{i=1}^{N}{(a_i-\mu)^2}}$ denote the mean and standard deviation of $x_{bn}$, respectively. $f_{BN}(x_{bn})$ is defined in Eq. (3).
$$f_{BN}(a_1,a_2,...,a_N)=(\gamma \frac{a_1-\mu}{\delta+\epsilon}+\beta,\gamma \frac{a_2-\mu}{\delta+\epsilon}+\beta,...,\gamma \frac{a_N-\mu}{\delta+\epsilon}+\beta)\eqno{(3)}$$
where, $\gamma\in \mathbb{R}^{+}$ and $\beta\in  \mathbb{R}$ are the parameters to be learned during training  and $\epsilon \textgreater 0$ is an arbitrarily small number. 

The batch normalization module eliminates the internal covariate shift and thus ensures a faster training process. Also, it regularizes the proposed model and enhances its local-feature extraction ability in supervised classification and unsupervised clustering. Meanwhile, different from the rectified linear unit ($ReLU$) that only considers positive numbers, $LeakReLU$ takes care of both positive and negative numbers, reducing the loss of features during the data transmission process. Detailed observations are shown in Section 4.3. The $LeakReLU$ activation is defined in Eq. (4).
\begin{equation}
	f_{LeakReLU} (x_{actv}) = \begin{cases}
	\quad \alpha  x_{actv}, & x_{actv} \textless 0 \\
	\quad x_{actv}, & x_{actv}  \geq 0	
		   \end{cases}
		   \tag{4}
\end{equation}
where, $x_{actv}$ is the input of the $LeakReLU$ unit and $\alpha$ is a coefficient for negative numbers. Following the widely recognized YOLOv3  \cite{F46}, we set $\alpha = 0.1$ in this paper.
 
\subsection{LSTM-based attention Layer}
As aforementioned, LSTMaN is proposed for relation extraction, including two LSTM-based attention layers. The two layers have the same structure, as shown in Fig. 2, where 'MatMul' is a matrix multiplication operation. The first layer is used to extract basic relationships from the input, while the second layer is responsible for mining the intricate connections among them. By extending the details of the relationships obtained before, the second layer helps to extract more complex regularizations hidden in data than the first layer. 

\begin{figure}[h]
  \centering
\includegraphics[width=8cm]{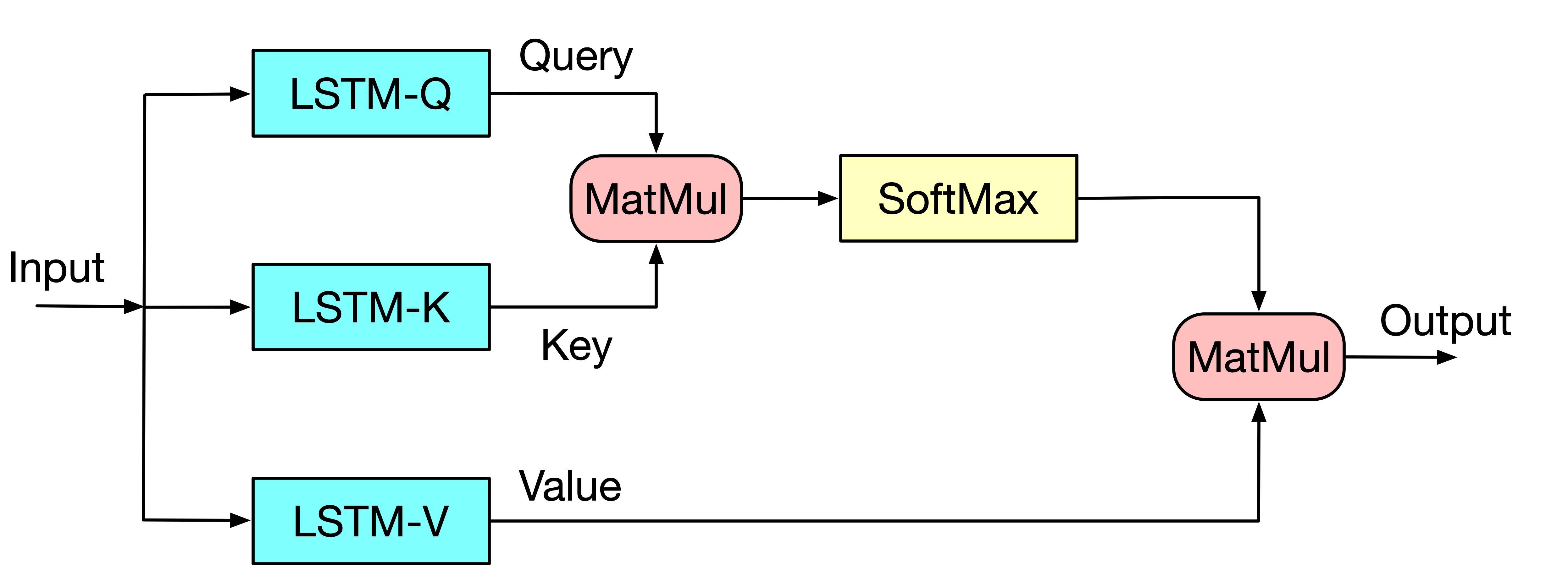}
\caption{Structure of the LSTM-based attention layer.}
\end{figure}
Unlike the existing models that hybridize attention and LSTM networks (see Section 2.2), an LSTM-based attention layer incorporates LSTM networks into an attention structure. In this network, a temporal query and a set of key-value pairs are mapped to an output, where Query, Key, and Value are matrices obtained from the feature extraction by the LSTM networks. The output is defined as a weighted sum of the values with sufficient representations. In this paper, each value is obtained by a compatibility function that mines the hidden relationships between a query and its corresponding key that already carry basic features. The query, key, and value matrices output by the three LSTM networks, $I_q$, $I_k$, $I_v$, are defined in Eqs. (5), (6) and (7), respectively.
$$ I_q = f_{LSTM-Q}(x) \eqno{(5)}$$
$$ I_k = f_{LSTM-K}(x) \eqno{(6)}$$
$$ I_v = f_{LSTM-V}(x) \eqno{(7)}$$
where, $x$ is the input of the layer and $f_{LSTM-Q}$, $f_{LSTM-K}$ and $f_{LSTM-V}$ are the LSTM functions for obtaining the query, key, and value matrices, respectively. 

Note that each of the three LSTM networks involves the same computational procedure at time step $t$ as a traditional LSTM network \cite{F72}. The following describes the computation operations in an LSTM network. Let $x_t$ and $h_t$ be the input vector and the hidden state vector of the LSTM network at $t$, respectively. Let $g_{t}^{u}$, $g_{t}^{f}$, $g_{t}^{o}$, and $g_{t}^{c}$ are the activation vectors of the input, forget, output, and cell state gates at $t$, respectively. Denote the elementwise multiplication by $\odot$. Let $\sigma$ and $tanh$ denote the logistic sigmoid and hyperbolic tangent functions, respectively. Let $W_{ux}$, $W_{fx}$, $W_{ox}$, and $W_{cx}$ be the weight matrices of $x_{t}$ at gates $g_{t}^{u}$, $g_{t}^{f}$, $g_{t}^{o}$, and $g_{t}^{c}$, respectively. Let $W_{uh}$, $W_{fh}$, $W_{oh}$, and $W_{ch}$ denote the weight matrices of $h_{t}$ at gates $g_{t}^{u}$, $g_{t}^{f}$, $g_{t}^{o}$, and $g_{t}^{c}$, respectively. Let $b_{u}$, $b_{f}$, $b_{o}$, and $b_{c}$ be the bias matrices of $h_{t}$ at gates $g_{t}^{u}$, $g_{t}^{f}$, $g_{t}^{o}$, and $g_{t}^{c}$, respectively. $g_{t}^{u}$, $g_{t}^{f}$, $g_{t}^{o}$, $g_{t}^{c}$, and $h_t$ are defined in Eqs. (8)-(12), respectively.
$$g_{t}^{u}=\sigma(W_{ux}x_t+W_{uh}h_{t-1}+b_{u})\eqno{(8)}$$
$$g_{t}^{f}=\sigma(W_{fx}x_t+W_{fh}h_{t-1}+b_{f})\eqno{(9)}$$
$$g_{t}^{o}=\sigma(W_{ox}x_t+W_{oh}h_{t-1}+b_{o})\eqno{(10)}$$
$$g_{t}^{c}=  g_{t}^{f}\odot g_{t-1}^{c} + g_{t}^{u} \odot tanh(W_{cx}x_t+W_{ch}h_{t-1}+b_{c})\eqno{(11)}$$
$$h_t=g_{t}^{o} \odot tanh(g_{t}^{c}) \eqno{(12)}$$

After $x$ goes through the LSTM networks, $I_q$, $I_k$, and $I_v$ carry sufficient long- and short-term features. They are then fed into the attention structure, and its output matrix, $ O_{Att}$, is defined in Eq. (13).
$$ O_{Att} = f_{SoftMax}(I_q \cdot  I_{k}^{T}) \cdot I_{v}  \eqno{(13)}$$
where, $f_{SoftMax}$ is a commonly used function to compute the possibilities of a certain matrix, and $I_{k}^{T}$ is the transpose of $I_{k}$. 

Note that we concatenate the local features obtained by TFN and the global relationships obtained by LSTMaN in  RTFN. Let $O_{TFN}$ and $O_{LSTMaN}$ denote the output matrices of TFN and LSTMaN, respectively. The output of RTFN, $O_{RTFN}$, is defined in Eq. (14).
$$O_{RTFN}=f_{concat}([O_{TFN},O_{LSTMaN}])\eqno{(14)}$$
where, $f_{concat}$ is the CONCAT function.
\begin{figure}[h]
  \centering
\includegraphics[width=8cm]{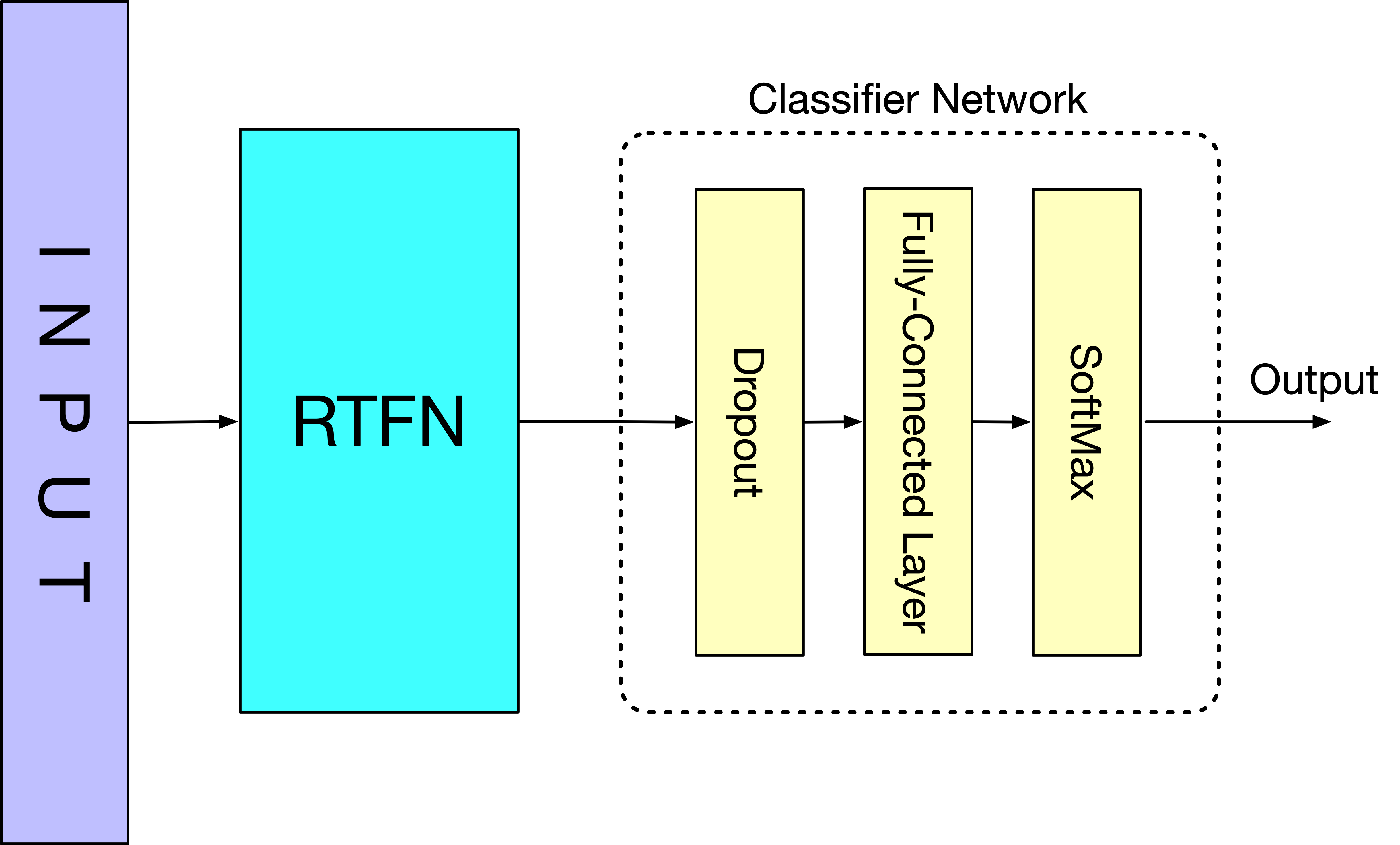}
\caption{RTFN-based supervised structure.}
\end{figure}

\subsection{RTFN-based Supervised Structure}

The RTFN-based supervised structure is shown in Fig. 3, where a dropout layer and a fully-connected layer are cascaded to the output of RTFN. To be specific, we introduce the dropout layer to avoid overfitting during the training process. The fully-connected layer performs as the classifier. That we simply use the dropout and fully-connected layers is because the features extracted by RTFN are sufficiently good, and thus a complicated classifier network is not necessary. Like other commonly used supervised algorithms \cite{F12,F14,F26,F31}, we use the cross-entropy function to compute the average difference between the ground truth labels and their corresponding prediction results, $\mathcal{L}_{super}$, as written in Eq. (15).

$$ \mathcal{L}_{super} =  - \frac {1}{n} \sum_{i=1}^{n} (\widehat{Y}_{i}^{train} log(p_{i})) \eqno{(15)}$$
where, $n$ is the number of samples, and $\widehat{Y}_{i}^{train}$ and $p_{i}$, $i=1,2,...,n$, are the ground truth label of the $i$-th sample and its corresponding prediction output, respectively.

\begin{figure}[h]
  \centering
\includegraphics[width=10cm]{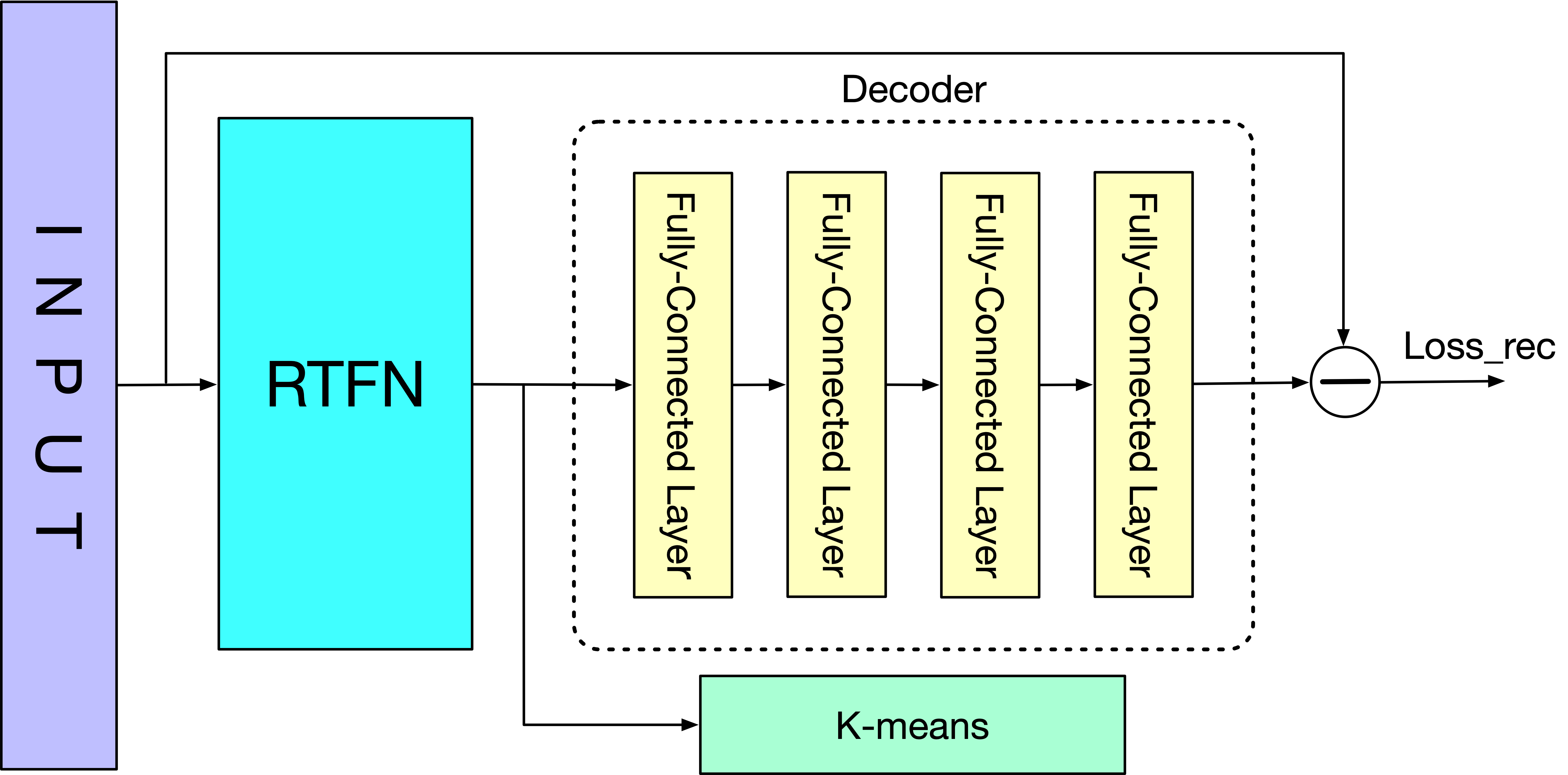}
\caption{RTFN-based unsupervised clustering.}
\end{figure}

\subsection{RTFN-based Unsupervised Clustering}
The RTFN-based unsupervised clustering is based on a widely adopted auto-encoder unsupervised structure, as shown in Fig. 4. It is primarily composed of an RTFN-based encoder, a decoder, and a K-means algorithm \cite{F47}. RTFN is responsible for obtaining as many useful representations from the input as possible. The decoder is made up of four fully-connected layers, helping to reconstruct the features captured by RTFN. Besides, the K-means algorithm acts as the unsupervised classifier.

Different from taking the k-means loss into account \cite{F49,F15,F48}, the RTFN-based unsupervised clustering only depends on the reconstruction loss (i.e., the mean square error), $ \mathcal{L}_{rec}$, as defined in Eq. (16).

$$  \mathcal{L}_{rec} =   \frac {1}{n} \sum_{i=1}^{n} (\breve{X}_{i}^{train} - X_{i}^{rec})^{2} \eqno{(16)}$$
where $\breve{X}_{i}^{train}$ and $ X_{i}^{rec}$, $i=1,2,...,n$, are the input and the decoder output of the $i$-th sample, respectively.

No matter the supervised or unsupervised structure, our goal is to minimize its loss function, $\mathcal{L}$, by finding the optimal parameters, $\theta^{*}$, where $\mathcal{L}(\theta^{*})$ is infinitely close to 0. This paper uses the gradient-descent method to approximate $\theta^{*}$ for our proposed structure. Let $\theta_{t}$ and $l_{rate}$ denote the parameters and the learning rate at the $t$-th training epoch, respectively. $\theta_{t}$ is updated by Eq. (17).
$$\theta_t = \theta_{t-1} - l_{rate}  \nabla_{\theta_{t-1}} \mathcal{L}(\theta_{t-1})  \eqno{(17)}$$
where, $\nabla_{\theta_{t-1}}$ represents the gradient at the $(t-1)$-th training epoch.
\section{Experiments and Analysis}
\label{others}

This section first introduces the experimental setup and performance metrics and then focus on the ablation study. Finally, the RTFN-based supervised structure and unsupervised clustering are evaluated, respectively.

\begin{table}[p]
  \caption{Details of 85 univariate time series datasets. Those marked with 'YES' are also used for unsupervised clustering experiments.}
  \label{sample-table}
  \centering
  \Huge
  \resizebox{72mm}{94mm}{
\begin{tabular}{|c|c|c|c|c|c|c|}
\toprule
Dataset                        & TrainSize & TestSize & Classes    & SeriesLength & Type               & Unsupervised \\\hline
Adiac                          & 390       & 391      & 37         & 176 	& Image              &         \\
ArrowHead                      & 36        & 175      & 3          & 251          & Image              &    YES          \\
Beef                           & 30        & 30       & 5          & 470          & Spectro            & YES          \\
BeetleFly                      & 20        & 20       & 2          & 512          & Image              & YES          \\
BirdChicken                    & 20        & 20       & 2          & 512          & Image              & YES          \\
Car                            & 60        & 60       & 4          & 577          & Sensor             & YES          \\
CBF                            & 30        & 900      & 3          & 128          & Simulated &     \\
ChlorineConcentration          & 467       & 3840     & 3          & 166          & Sensor             & YES \\
CinCECGTorso                   & 40        & 1380     & 4          & 1639         & Sensor             &              \\
Coffee                         & 28        & 28       & 2          & 286          & Spectro            & YES          \\
CricketX                       & 390       & 390      & 12         & 300          & Motion             &              \\
CricketY                       & 390       & 390      & 12         & 300          & Motion             &              \\
CricketZ                       & 390       & 390      & 12         & 300 & Motion             &              \\
DiatomSizeReduction            & 16        & 306      & 4          & 345          & Image              & YES          \\
DistalPhalanxOutlineAgeGroup   & 400       & 139      & 3          & 80           & Image              & YES          \\
DistalPhalanxOutlineCorrect    & 600       & 276      & 2          & 80           & Image              & YES          \\
Earthquakes                    & 322       & 139      & 2          & 512          & Sensor             &              \\
ECG200                         & 100       & 100      & 2          & 96           & ECG                & YES          \\
ECG5000                        & 500       & 4500     & 5          & 140          & ECG                &              \\
ECGFiveDays                    & 23        & 861      & 2          & 136          & ECG                & YES          \\
FaceFour                       & 24        & 88       & 4          & 350          & Image              &              \\
FacesUCR                       & 200       & 2050     & 14         & 131          & Image              &              \\
FordA                          & 3601      & 1320     & 2          & 500          & Sensor             &              \\
FordB                          & 3636      & 810      & 2          & 500          & Sensor             &              \\
GunPoint                       & 50        & 150      & 2          & 150          & Motion             & YES          \\
Ham                            & 109       & 105      & 2          & 431          & Spectro            & YES          \\
HandOutlines                   & 1000      & 370      & 2          & 2709         & Image              &     \\
Haptics                        & 155       & 308      & 5          & 1092         & Motion             &              \\
Herring                        & 64        & 64       & 2          & 512          & Image              & YES          \\
InlineSkate                    & 100       & 550      & 7          & 1882         & Motion             &              \\
InsectWingbeatSound            & 220       & 1980     & 11         & 256          & Sensor             &              \\
ItalyPowerDemand               & 67        & 1029     & 2          & 24           & Sensor             &     \\
Lightning2                     & 60        & 61       & 2          & 637 & Sensor             & YES          \\
Lightning7                     & 70        & 73       & 7          & 319          & Sensor             &              \\
Mallat                         & 55        & 2345     & 8          & 1024         & Simulated          &              \\
Meat                           & 60        & 60       & 3          & 448          & Spectro            & YES          \\
MedicalImages                  & 381       & 760      & 10         & 99           & Image              &              \\
MiddlePhalanxOutlineAgeGroup   & 400       & 154      & 3          & 80           & Image              & YES          \\
MiddlePhalanxOutlineCorrect    & 600       & 291      & 2          & 80           & Image              & YES          \\
MiddlePhalanxTW                & 399       & 154      & 6          & 80           & Image              & YES          \\
MoteStrain                     & 20        & 1252     & 2          & 84           & Sensor             & YES          \\
OliveOil                       & 30        & 30       & 4          & 570          & Spectro            &              \\
OSULeaf                        & 200       & 242      & 6          & 427          & Image              & YES          \\
Plane                          & 105       & 105      & 7          & 144          & Sensor             & YES          \\
ProximalPhalanxOutlineAgeGroup & 400       & 205      & 3          & 80           & Image              & YES          \\
ProximalPhalanxOutlineCorrect  & 600       & 291      & 2          & 80           & Image              &              \\
ProximalPhalanxTW              & 400       & 205      & 6          & 80           & Image              & YES          \\
ShapeletSim                    & 20        & 180      & 2          & 500          & Simulated          &              \\
ShapesAll                      & 600       & 600      & 60         & 512          & Image              &              \\
SonyAIBORobotSurface1          & 20        & 601      & 2          & 70           & Sensor             & YES          \\
SonyAIBORobotSurface2          & 27        & 953      & 2          & 65           & Sensor             & YES          \\
Strawberry                     & 613       & 370      & 2          & 235          & Spectro            &              \\
SwedishLeaf                    & 500       & 625      & 15         & 128          & Image              & YES          \\
Symbols                        & 25        & 995      & 6          & 398          & Image              & YES          \\
SyntheticControl               & 300       & 300      & 6          & 60           & Simulated          &              \\
ToeSegmentation1               & 40        & 228      & 2          & 277          & Motion             & YES          \\
ToeSegmentation2               & 36        & 130      & 2          & 343          & Motion             & YES          \\
Trace                          & 100       & 100      & 4          & 275          & Sensor             &              \\
TwoPatterns                    & 1000      & 4000     & 4          & 128          & Simulated          & YES          \\
TwoLeadECG                     & 23        & 1139     & 2          & 82           & ECG                & YES          \\
UWaveGestureLibraryAll         & 896       & 3582     & 8          & 945          & Motion             &              \\
UWaveGestureLibraryX           & 896       & 3582     & 8          & 315          & Motion             &              \\
UWaveGestureLibraryY           & 896       & 3582     & 8          & 315          & Motion             &              \\
UWaveGestureLibraryZ           & 896       & 3582     & 8          & 315          & Motion             &              \\
Wafer                          & 1000      & 6164     & 2          & 152          & Sensor             & YES          \\
Wine                           & 57        & 54       & 2          & 234          & Spectro            & YES          \\
WordSynonyms                   & 267       & 638      & 25         & 270          & Image              & YES          \\
ACSF1                          & 100       & 100      & 10         & 1460         & Device             &              \\
BME                            & 30        & 150      & 3          & 128          & Simulated          &              \\
Chinatown                      & 20        & 345      & 2          & 24           & Traffic            &              \\
Crop                           & 7200      & 16800    & 24         & 46           & Image              &              \\
DodgerLoopDay                  & 78        & 80       & 7          & 288          & Sensor             &              \\
DodgerLoopGame                 & 20        & 138      & 2          & 288          & Sensor             &              \\
DodgerLoopWeekend              & 20        & 138      & 2          & 288          & Sensor             &              \\
GunPointAgeSpan                & 135       & 316      & 2          & 150          & Motion             &              \\
GunPointMaleVersusFemale       & 135       & 316      & 2          & 150          & Motion             &              \\
GunPointOldVersusYoung         & 136       & 315      & 2          & 150          & Motion             &              \\
InsectEPGRegularTrain          & 62        & 249      & 3          & 601          & EPG                &              \\
InsectEPGSmallTrain            & 17        & 249      & 3          & 601          & EPG                &              \\
MelbournePedestrian            & 1194      & 2439     & 10         & 24           & Traffic            &              \\
PowerCons                      & 180       & 180      & 2          & 144          & Power              &              \\
Rock                           & 20        & 50       & 4          & 2844         & Spectrum           &              \\
SemgHandGenderCh2              & 300       & 600      & 2          & 1500         & Spectrum           &              \\
SemgHandMovementCh2            & 450       & 450      & 6          & 1500         & Spectrum           &              \\
SemgHandSubjectCh2             & 450       & 450      & 5          & 1500         & Spectrum           &              \\
SmoothSubspace                 & 150       & 150      & 3          & 15           & Simulated          &              \\
UMD                            & 36        & 144      & 3          & 150          & Simulated          &             

\\

\bottomrule
   
\end{tabular}

}
\end{table}

\begin{table}[!htb]
  \caption{Details of 30 multivariate time series datasets. Abbreviations: AS - Audio Spectra, ECG - Electrocardiogram, EEG - Electroencephalogram, HAR - Human Activity Recognition, MEG - Magnetoencephalography.}
  \label{sample-table}
  \centering
  \resizebox{100mm}{50mm}{
\label {dd}
\begin{tabular}{|c|c|c|c|c|c|c|c|}
\toprule
Index& Dataset                 & TrainSize & TestSize & NumDimensions & SeriesLength & Classes & Type    \\
     \hline
AWR  & ArticularWordREcognition & 275       & 300      & 9             & 144          & 25      & Motion  \\
AF   & AtrialFibrillation       & 15        & 15       & 2             & 640          & 3       & ECG     \\
BM   & BasicMotions             & 40        & 40       & 6             & 100          & 4       & HAR     \\
CT   & CharacterTrajectories    & 1422      & 1436     & 3             & 182          & 20      & Motion  \\
CR   & Critcket                 & 108       & 72       & 6             & 1197         & 12      & HAR     \\
DDG  & DuckDuckGeese            & 50        & 50       & 1345          & 270          & 5       & AS      \\
EW   & EigenWorm                & 128       & 131      & 6             & 17894        & 4       & Motion  \\
EP   & Epliyepsy                & 137       & 138      & 3             & 206          & 4       & HAR     \\
EC   & EthanolConcentration     & 261       & 263      & 3             & 1751         & 4       & HAR     \\
ER   & ERing                    & 30        & 270      & 4             & 65           & 6       & Other   \\
FD   & FaceDetection            & 5890      & 3524     & 144           & 62           & 2       & EEG/MEG \\
FM   & FingerMovements          & 316       & 100      & 28            & 50           & 2       & EEG/MEG \\
HMD  & HandMovementDirection    & 160       & 74       & 10            & 400          & 4       & EEG/MEG \\
HW   & Handwriting              & 150       & 850      & 3             & 152          & 26      & HAR     \\
HB   & Heartbeat                & 204       & 205      & 61            & 405          & 2       & AS      \\
IW   & InsectionWingbeat        & 30000     & 20000    & 200           & 30           & 10      & AS      \\
JV   & JapaneseVowels           & 270       & 370      & 12            & 29           & 9       & AS      \\
LIB  & Libras                   & 180       & 180      & 2             & 45           & 15      & HAR     \\
LSST & LSST                     & 2459      & 2466     & 6             & 36           & 14      & Other   \\
MI   & MotorImagery             & 278       & 100      & 64            & 3000         & 2       & EEG/MEG \\
NATO & NATOPS                   & 180       & 180      & 24            & 51           & 6       & HAR     \\
PD   & PenDigits                & 7494      & 3498     & 2             & 8            & 10      & EEG/MEG \\
PEMS & PEMSF                    & 267       & 173      & 963           & 144          & 7       & EEG/MEG \\
PS   & Phoneme                  & 3315      & 3353     & 11            & 217          & 39      & AS      \\
RS   & RacketSports             & 151       & 152      & 6             & 30           & 4       & HAR     \\
SRS1 & SelfRegulationSCP1       & 268       & 293      & 6             & 896          & 2       & EEG/MEG \\
SRS2 & SelfRegulationSCP2       & 200       & 180      & 7             & 1152         & 2       & EEG/MEG \\
SAD  & SpokenArbicDigits        & 6599      & 2199     & 13            & 93           & 10      & AS      \\
SWJ  & StandWalkJump            & 12        & 15       & 4             & 2500         & 3       & ECG     \\
UW   & UWaveGuestureLibrary     & 120       & 320      & 3             & 315          & 8       & HAR    \\
\bottomrule
\end{tabular}
  }
\end{table}

\subsection{Experimental Setup}

Extensive experiments in supervised classification and unsupervised clustering have been conducted. This section introduces the standard datasets first and the implementation details later.
 
\textbf{Supervised Classification Datasets.}  A univariate time series refers to a series of time-ordered data points associated with a time-dependent variable. Such a sequence contains local and global patterns of data. Local patterns show significant changes in the data, and global patterns reflect the overall trend of data \cite{F7}. A multivariate time series consists of multiple concurrent univariate time series, each associated with a time-dependent variable. Like univariate time series, a multivariate time series also contain local and global pattern information. Besides, as each variable has some dependency on other variables, a multivariate time series contains relationship information between variables \cite{F4}. In the supervised area, time series data are labeled. A supervised algorithm learns the characteristics of the input data and maps the data to labels. For a univariate time series, a supervised algorithm needs to mine the local and global patterns from the data, such as EE \cite{F16}, COTE \cite{F14}, LPS \cite{F61}, LCE \cite{F18}, ConvTime \cite{F23}, ResNet-Transformer models \cite{F26}, etc. For a multivariate time series, a supervised algorithm focuses on not only the local and global patterns of each variable but also the relationships between variables, e.g., HUML \cite{F20}, SMTS \cite{F76}, mv-ARF \cite{F60}, WEASEL + MUSE \cite{F56}, TapNet \cite{F58}, FCN-MLSTM \cite{F31}, and so on.

We evaluate the performance of the RTFN-based supervised structure by a number of univariate and multivariate time series datasets. For the univariate time series, UCR2018 \cite{F7} is one of the authoritative data archives, which contains 128 datasets with different lengths in a variety of application areas. We select 85 standard datasets from the UCR2018 archive, consisting of 65 'short-medium' and 20 'long' time series datasets. In this paper, a 'long' dataset is a dataset with a length of over 500. The details of these datasets are shown in Table 1. As for the multivariate time series, UEA2018 \cite{F4} is a commonly used data archive, including 30 datasets in seven application areas, i.e., audio spectra, electrocardiogram, electroencephalogram, human activity recognition, motion, eagnetoencephalography and other. Their details are listed in Table 2. 

\textbf{Unsupervised Clustering Datasets.} In the unsupervised area, time series data are not labeled. An unsupervised algorithm discovers the previously undetected patterns in a dataset. For a univariate time series, an unsupervised algorithm aims to learn the underlying structure or distribution in the data, such as K-means \cite{F47}, UDFS \cite{F62}, K-shape \cite{F68}, DTCR \cite{F15}, IDEC \cite{F49}, etc. For a multivariate time series, an unsupervised algorithm learns not only the underlying structure or distribution of each variable but also the relationships between variables, e.g., USAD \cite{F77}.

Following the protocol used in \cite{F49,F15,F48}, we verify the performance of the RTFN-based unsupervised clustering by 36 standard datasets selected from the UCR2018 archive. They are marked with 'YES' in column 'Unsupervised' in Table 1.

\textbf{Implementation Details.} Firstly, we introduce the parameter settings for TFN. As mentioned in Section 3.2, each Conv1D block contains a 1-dimensional CNN module. In the Conv1D block directly connected to the residual junction, the 1-dimensional CNN module also has 128 channels, each with a kernel size of 1. In each multi-head Conv1D layers, each of the four 1-dimensional CNN modules has 32 channels. Motivated by InceptionTime \cite{F24}, we set the kernel sizes of the four CNN modules to 5, 8, 11, and 17, respectively. Following the previous works in \cite{F30,F29,F26,F46}, we set the decay value of the batch normalization module to 0.9, which helps to accelerate the training by reducing the internal covariate shift of time series data.

In the two Conv1D blocks next to the input of RTFN, each 1-dimensional CNN module has 128 channels, each with a kernel size of 11. The following explains why 11 is chosen. As references \cite{F24,F26,F27,F23,F25,F33} suggest, we choose 3, 5, 7, 9, 11, and 13 as the candidate kernel sizes for the two ConvlD blocks above. We select eight univariate and four multivariate datasets from the UCR2018 and UEA2018 archives, respectively. The eight univariate datasets contain four ‘short-medium’ and four ‘long’ time series datasets. Table 3 shows the top-1 accuracy results with six different kernel sizes used in the two Conv1D blocks. A larger kernel size usually leads to higher accuracies because such a kernel has a broader receptive field and thus captures richer local features. Clearly, kernel sizes 11 and 13 result in better performance than 3, 5, 7, and 9. Kernel sizes 11 and 13 lead to similar accuracy results on each dataset. On the other hand, a larger kernel size means the corresponding convolutional operations consume more computing resources. To compromise between accuracy and complexity, we set the kernel size of the two Conv1D blocks to 11.

\begin{table}[!htb]
  \caption{Results of the top-1 accuracy vs. the kernel size.}
  \label{sample-table}
  \centering
  \resizebox{120mm}{20mm}{
\begin{tabular}{|c|ccccccc|}
\hline
Type                                                              & Dataset             & Kernel size = 3        & Kernel size = 5        & Kernel size = 7              & Kernel size = 9                & Kernel size = 11                & Kernel size = 13                \\ \hline
\multicolumn{1}{|c|}{\multirow{9}{*}{Univariate   Time Series}}   & ECG200              & 0.9      & 0.91     & 0.91           & 0.92             & 0.92              & \textbf{0.93}     \\
\multicolumn{1}{|c|}{}                                            & Lighting7           & 0.863014 & 0.876712 & 0.890411       & \textbf{0.90411} & \textbf{0.90411}  & \textbf{0.90411}  \\
\multicolumn{1}{|c|}{}                                            & Ham                 & 0.761905 & 0.761905 & 0.780952       & 0.780952         & \textbf{0.809524} & \textbf{0.809524} \\
\multicolumn{1}{|c|}{}                                            & Wine                & 0.87037  & 0.87037  & 0.888889       & 0.888889         & \textbf{0.907407} & \textbf{0.907407} \\
\multicolumn{1}{|c|}{}                                            & SemgHandGenderCh2   & 0.876667 & 0.96     & 0.91           & 0.913333         & \textbf{0.923333} & \textbf{0.923333} \\
\multicolumn{1}{|c|}{}                                            & SemgHandMovementCh2 & 0.6      & 0.616667 & 0.66667        & 0.716667         & \textbf{0.757778} & \textbf{0.757778} \\
\multicolumn{1}{|c|}{}                                            & SemgHandSubjectCh2  & 0.788889 & 0.816667 & 0.833333       & 0.85             & \textbf{0.897778} & \textbf{0.897778} \\
\multicolumn{1}{|c|}{}                                            & Rock                & 0.84     & 0.84     & 0.84           & 0.86             & \textbf{0.88}     & \textbf{0.88}     \\ \cline{2-8} 
\multicolumn{1}{|c|}{}                                            & MeanACC             & 0.812606 & 0.831540 & 0.840032       & 0.854244         & 0.874991          & \textbf{0.876241} \\ \hline
\multicolumn{1}{|c|}{\multirow{5}{*}{Multivariate   Time Series}} & AF                  & 0.467    & 0.467    & \textbf{0.533} & \textbf{0.533}   & \textbf{0.533}    & \textbf{0.533}    \\
\multicolumn{1}{|c|}{}                                            & FD                  & 0.629    & 0.631    & 0.649          & 0.656            & \textbf{0.67}     & \textbf{0.67}     \\
\multicolumn{1}{|c|}{}                                            & HMD                 & 0.649    & 0.649    & \textbf{0.662} & \textbf{0.662}   & \textbf{0.662}    & \textbf{0.662}    \\
\multicolumn{1}{|c|}{}                                            & HB                  & 0.727    & 0.748    & 0.751          & 0.751            & \textbf{0.785}    & \textbf{0.785}    \\ \cline{2-8} 
\multicolumn{1}{|c|}{}                                            & MeanACC             & 0.618    & 0.624    & 0.649          & 0.651            & \textbf{0.663}    & \textbf{0.663}    \\ \hline
\end{tabular}
}
\end{table}

Secondly, we introduce the parameter settings for LSTMaN. As described in Section 3.3, there are two LSTM-based attention layers. In each layer, as references \cite{F27,F31} suggest, we set the number of hidden units in each LSTM network to 128. 

Last but not least, we dynamically adjust the learning rate during the training process. Let the total number of training epochs and the size of each decay period denoted by $N_{tot}$ and $N_{dec}$, respectively. Let $l_{rate}(j), j=1,2, ..., J$, denote the learning rate of the $j-th$ decay period, where $J = \lceil N_{tot}/N_{dec} \rceil - 1$. Its definition is written in Eq. (18).
 $$ l_{rate}(j) = (1-d_{rate}) \times l_{rate}(j-1)  \eqno{(18)}$$
where $d_{rate}$ and $J$ are the decay rate of $l_{rate}$ and the total number of the decay periods, respectively. In this paper, we set $l_{rate}(0)$ = 0.01 and $d_{rate}$ = 0.1. Once $l_{rate}$ is smaller than 0.0001, we fix it to 0.0001. The RMSPropOptimizer of Tensorflow \footnote{\url{https://tensorflow.google.cn/}} is used to tune the parameters of our proposed RTFN structures for supervised classification and unsupervised clustering. According to \cite{F75}, we set the RMSPropOptimizer's momentum term to 0.9 to avoid falling into local minima during  training. Besides, we use the dropout layer and $L_2$ regularization to avoid overfitting during the training process. As references \cite{F30,F34, F42, F41,F43} suggest, the dropout layer's ratio value is set to 0.5.

All experiments are run on a computer with Ubuntu 18.04 OS, an Nvidia GTX 1070Ti GPU with 8GB, an Nvidia GTX 1080Ti GPU with 11GB, and an AMD R5 1400 CPU with 16G RAM.

\subsection{Performance Metrics}
To evaluate the performance of various algorithms in terms of supervised classification and unsupervised clustering, we adopt a number of well-known performance metrics explained below.

\textbf{Supervised Classification.} Three metrics are used to rank different supervised algorithms in terms of the top-1 accuracy, including ‘win’/‘tie’/‘lose’, mean accuracy (MeanACC), and AVG\_rank. To be specific, for an arbitrary algorithm, its ‘win’, ‘tie’, and ‘lose’ values indicate on how many datasets this algorithm performs better than, equivalent to, and worse than the others, respectively. For each algorithm, the ‘best’ value is the summation of its corresponding ‘win’ and ‘tie’ values, while the ‘total’ value is the total number of datasets tested. In addition, the AVG\_rank score measures the average difference between the accuracy values of a model and the best accuracy values among all models \cite{F14,F18,F19,F26}.

\textbf{Unsupervised Clustering.} Note that the top-1 accuracy is not applicable to unsupervised clustering. Instead, we use a widely adopted performance indicator, the rand index (RI) \cite{F50}, $RI$, as defined in Eq. (19).
$$ RI = \frac {PTP + NTP}{s(s-1)/2}  \eqno{(19)}$$
where $PTP$ and $NTP$ are the numbers of the positive and negative time series pairs in the clustering, respectively, and $s$ is the dataset size. Besides, we denote the average RI value of a certain algorithm by ‘AVG RI’.

\subsection{Ablation Study}
As shown in Fig. 1, RTFN mainly consists of a temporal feature network for local-feature extraction, i.e., TFN, and an LSTM-based attention network for relation extraction, i.e., LSTMaN. 

\begin{table}[!htb]
  \caption{The top-1 accuracy results of different supervised algorithms on 12 selected datasets.}
  \label{sample-table}
  \centering
  \Huge
  \resizebox{120mm}{20mm}{
\begin{tabular}{|c|cc|ccc|ccccc|}
\toprule
Type                                                                                      & Dataset&SeriesLength             & TFN  & TFN w $ReLU$   & TFN  w/o $SelAtt$  & TFN+1LSTMaL & TFN+2LSTMaL & TFN+3LSTMaL & TFN+AttLSTM & TFN+LSTMTAG             \\\hline
\multirow{9}{*}{\begin{tabular}[c]{@{}l@{}}Univariate  Time\\  Series\end{tabular}} & ECG200 &96             & 0.84          & 0.82          & 0.8     & 0.9       & \textbf{0.92}        & \textbf{0.92}     & 0.91             & 0.89                  \\
                                                                                          & Lighting7    & 319      & 0.821918    & 0.808219     &0.808219    & 0.849315      & \textbf{0.90411}    & \textbf{0.90411}  & 0.849315         & 0.849315           \\
                                                                                          & Ham           &431      &  0.714286    & 0.619048     & 0.619048  & 0.761905       & \textbf{0.809524}    & \textbf{0.809524} & 0.780952         & 0.761905         \\
                                                                                          & Wine           & 234    & 0.833333    &  0.611111   &   0.555556  & 0.888889       & \textbf{0.907407}   & \textbf{0.907407} & 0.888889         & 0.87037           \\
                                                                                          & SemgHandGenderCh2 &1500  &  0.796667     &  0.783333    &  0.651667 &  0.866667  & \textbf{0.923333}      & \textbf{0.923333} & 0.91             & 0.84              \\
                                                                                          & SemgHandMovementCh2 &1500&   0.595555       &0.56      & 0.513333  & 0.611111       & \textbf{0.757778}   & \textbf{0.757778} & 0.56             & 0.56              \\
                                                                                          & SemgHandSubjectCh2 &1500 &  0.793333    &0.788889          & 0.74  & 0.8         & \textbf{0.897778}      & \textbf{0.897778} & 0.873333         & 0.74              \\ 
                                                                                          & Rock             &2844   &   0.7      & 0.64         &  0.62       & 0.82          & \textbf{0.88}       & \textbf{0.88}     & 0.86             & 0.68               \\ \cmidrule{2-11}
                                                                                          & \multicolumn{2}{c|}{MeanACC}       &  0.761887     &   0.703825   &0.663478  & 0.812236        & \textbf{0.874991}   & \textbf{0.874991} & 0.829061         & 0.773949         \\\hline
\multirow{5}{*}{\begin{tabular}[c]{@{}l@{}}Multivariate  Time \\ Series\end{tabular}}     & AF      &640            &   0.4       &    0.333     &  0.267     & 0.467           & \textbf{0.533}   & \textbf{0.533}    & 0.467            & 0.2                 \\
                                                                                          & FD      &62            &  0.555       &  0.545       &  0.519    & 0.614        & \textbf{0.67}      & \textbf{0.67}     & 0.631            & 0.555                \\
                                                                                          & HMD     &400            &  0.544       &    0.481       &   0.5   & 0.649        & \textbf{0.662}      & \textbf{0.662}    & 0.649            & 0.649               \\
                                                                                          & HB          &405        &    0.547       &0.535         & 0.515   & 0.761       & \textbf{0.785}       & \textbf{0.785}    & 0.727            & 0.547               \\ \cmidrule{2-11}
                                                                                          & \multicolumn{2}{c|}{MeanACC}          &     0.512      &   0.474      &  0.450  & 0.623           & \textbf{0.663}    & \textbf{0.663}    & 0.619            & 0.488             \\
                                                                                          \bottomrule
\end{tabular}
}
\end{table}

\begin{table}[!htb]
  \caption{The RI results of different unsupervised algorithms on four selected datasets.}
  \label{sample-table}
  \centering
  \Huge
  \resizebox{100mm}{10mm}{

\begin{tabular}{|c|ccc|ccccc|}
\toprule
Dataset   & TFN    & TFN  w $ReLU$  & TFN  w/o $SelAtt$ & TFN+1LSTMaL & TFN+2LSTMaL & TFN+3LSTMaL & TFN+AttLSTM & TFN+LSTMTAG  \\ \hline
Beef      &      0.6267   &    0.5945    & 0.5402 & 0.7034         &\textbf{ 0.7655}    &\textbf{ 0.7655}            & 0.7057           & 0.7034           \\
Car       &     0.6418    &   0.6354      &0.626  & 0.6708            & \textbf{0.7169} & 0.7028            & 0.6898           & 0.6418           \\
ECG200    &    0.6533     &   0.6315     & 0.6018 & 0.7018        &\textbf{ 0.7285}    &\textbf{ 0.7285}            & 0.7018           & 0.7018            \\
Lighting2 &     0.5373    &    0.5119    & 0.4966 & 0.5729        & \textbf{0.6230}     &\textbf{ 0.6230}             & 0.5770           & 0.5770            \\\hline
AVG   RI  &    0.6148     &    0.5933    & 0.5662 & 0.6622         & \textbf{0.7085}    & 0.7050            & 0.6686           & 0.6560          \\
\bottomrule
\end{tabular}
}
\end{table}

\subsubsection{Temporal Feature Network}

TFN is featured with $LeakReLU$-based activation and self-attention. To study the effectiveness of the two components, we compare three TFN variants listed below. 

\begin{itemize}
\item[-] TFN: the proposed TFN, where $LeakReLU$ and self-attention are used.
\item[-] TFN w $ReLU$: TFN with $ReLU$ instead of $LeakReLU$.
\item[-] TFN w/o $SelAtt$: TFN without the self-attention layer.

\end{itemize}

For the performance comparison of supervised classification, we select 12 datasets from the UCR2018 and UEA2018 archives, including eight univariate and four multivariate datasets. These datasets are also used in Table 3. For the performance comparison of unsupervised clustering, we select four univariate datasets from the UCR2018 archive. 

The top-1 accuracy and RI results obtained by different supervised and unsupervised algorithms are shown in Tables 4-5, respectively. It is easily observed that TFN outperforms TFN w $ReLU$ on each dataset for supervised classification or unsupervised clustering. For example, the top-1 accuracy values of TFN and TFN w $ReLU$ are 0.833333 and 0.611111, respectively. Unlike $ReLU$ that focuses on positive numbers only, $LeakReLU$ makes use of both positive and negative numbers, helping to avoid the loss of the extracted features during their transformation. Thus, $LeakReLU$ can mine more local features from the input. That is why $LeakReLU$ improves both the supervised and unsupervised performance of TFN, compared with $ReLU$. We then compare TFN and TFN w/o $SelAtt$ in terms of the supervised classification and unsupervised clustering. TFN overweighs TFN w/o $SelAtt$ on all the datasets. That is because the $SelAtt$ layer can relate different positions of time series data, enriching the extracted features. So, embedding the $SelAtt$ layer in TFN helps to enhance its supervised and unsupervised performance. Therefore, $LeakReLU$ and self-attention are necessary for TFN.

\subsubsection{The LSTM-based Attention Network}

In RTFN, LSTMaN consists of two LSTM-based attention layers. To investigate the effectiveness of LSTMaN, we compare five RTFN structures with the following relation-extraction components.  

\begin{itemize}
\item[-] 1LSTMaL: one LSTM-based attention layer.
\item[-] 2LSTMaL: two LSTM-based attention layers, i.e. the proposed LSTMaN.
\item[-] 3LSTMaL: three LSTM-based attention layers.
\item[-] AttLSTM: a cascading attention-LSTM model, where attention and LSTM layers simply pile up \cite{F31}.
\item[-] LSTMTAG: an embedding attention-LSTM model, where a trend attention gate is embedded into an LSTM structure \cite{F32}.
\end{itemize}

To make a fair comparison, TFN is used in each RTFN structure as the local-feature extraction network. In other words, no matter for supervised classification or unsupervised clustering, the corresponding RTFN structures are exactly the same, except for their relation-extraction components.

Firstly, we study the impact of the number of LSTM-based attention layers on the performance of RTFN. Between TFN+2LSTMaL and TFN+1LSTMaL, the former always performs better than the latter. In the 2LSTMaL structure, the second layer unfolds the details of the relationships among the features captured by the first layer and hence can discover those complicated representations ignored before. That is why 2LSTMaL mines more intricate relationships hidden in the data than 1LSTMaL. If comparing TFN+2LSTMaL and TFN+3LSTMaL, one can see that the two achieve equivalent performance in almost all cases except dataset 'Car', as illustrated in Tables 4-5. The following explains why. In the 3LSTMaL structure, the third layer is supposed to further extend the details of the relationships among those features extracted by the first and second layers. However, all intrinsic details have been explicitly unveiled in the second layer. In this case, the third layer only acts as an information transmission layer. This layer not only may lead to loss of features during their transmission but also consumes additional computing resources, especially on complicated datasets. 2LSTMaL aims at striking a balance between accuracy and model complexity. To further support this, we show the model complexity comparison of different supervised algorithms on four long time series datasets in Table 6. It is easily seen that TFN+2LSTMaL has a lower model complexity than TFN+3LSTMaL, e.g., their CPU time on dataset 'SemgHandGendeCh2' is 32.421894s and 35.440211s, respectively.

\begin{table}[!htb]
  \caption{Computational complexity comparison of TFN+1LSTMaL, TFN+2LSTMaL and TFN+3LSTMaL in terms of the supervised classification. Abbreviations: M -- Measured in Millions, s -- Measured in Seconds.}
  \label{sample-table}
  \centering
  \Huge
  \resizebox{100mm}{20mm}{

\begin{tabular}{|cccccc|}
\toprule
Algorithm        & Dataset                              & Parameters   (M) & CPU   only (s) & With   GPU 1080Ti (s) & With   GPU 1070Ti (s) \\ \hline

TFN+1LSTMaL & \multirow{3}{*}{SemgHandGendeCh2}    & 2.546755         & 30.277891      & 1.506916              & 1.603434              \\
TFN+2LSTMaL             &                                      & 2.851651         & 32.421894      & 2.127678              & 2.410963              \\
TFN+3LSTMaL &                                      & 3.262915         & 35.440211      & 2.893432              & 3.129045              \\
\hline

TFN+1LSTMaL & \multirow{3}{*}{SemgHandMovementCh2} & 3.315015         & 21.771511      & 1.30519               & 1.537834              \\
TFN+2LSTMaL       &                                      & 3.620167         & 24.561916      & 1.865625              & 2.028537              \\
TFN+3LSTMaL &                                      & 4.031431         & 26.660317      & 2.345234              & 2.834242              \\

\hline

TFN+1LSTMaL & \multirow{3}{*}{SemgHandSubjectCh2}  & 3.12295          & 20.742758      & 1.303414              & 1.529351              \\
TFN+2LSTMaL         &                                      & 3.428038         & 24.695192      & 1.864818              & 2.040713              \\
TFN+3LSTMaL &                                      & 3.839302         & 26.723078      & 1.934696              & 2.783453              \\
\hline

TFN+1LSTMaL & \multirow{3}{*}{Rock}                & 4.747221         & 6.989958       & 1.293770              & 1.352977              \\
TFN+2LSTMaL           &                                      & 5.042245         & 8.887336       & 1.360002              & 1.370617             \\
TFN+3LSTMaL &                                      & 5.453509         & 10.771745      & 1.636234              & 2.103425              \\

\bottomrule
\end{tabular}
 }
\end{table}

Secondly, we investigate the effectiveness of the proposed LSTMaN by comparing it with two well-recognized models based on attention and LSTM. It can be seen that TFN+2LSTMaL beats TFN+AttLSTM and TFN+LSTMTAG on each dataset for supervised classification or unsupervised clustering. The following explains the reasons. On the one hand, AttLSTM lacks in-depth attention to the internal connections among the already extracted representations during their transmission, and thus insufficient features are mined from data. Meanwhile, LSTMTAG is able to concentrate on the local variations of periodical data due to the LSTM structure with TAG embedded. However, it is not sensitive to the global variations of non-periodical data, which is not beneficial to discover complex connections hidden in the data, especially when facing long univariate datasets and multivariate ones. For instance, for TFN+LSTMTAG, its top-1 accuracy values on datasets 'Rock' and 'AF' are only 0.68 and 0.2, respectively. On the other hand, compared with TFN, TFN+2LSTMaL always obtains a higher accuracy value on each dataset, no matter in the aspect of supervised classification or unsupervised clustering. It clearly demonstrates that LSTMaN plays a non-trivial role in performance improvement. That is because 2LSTMaL extracts those intricate representations that may be ignored by TFN. In other words, LSTMaN and TFN complement each other in RTFN.

\begin{table}[p]
  \caption{Results of different supervised algorithms on 85 selected datasets.}
  \label{sample-table}
  \Huge
  \centering
  \resizebox{120mm}{97mm}{
\begin{tabular}{|c|ccccccccccccccc|}
\toprule
Dataset                        & \multicolumn{2}{c}{ \begin{tabular}[c]{@{}c@{}}Existing \\SOTA \end{tabular}}   & \multicolumn{2}{c}{\begin{tabular}[c]{@{}c@{}}USRL-\\FordA\cite{F52} \end{tabular}}    & \begin{tabular}[c]{@{}c@{}}Inception-\\Time\cite{F24} \end{tabular}   & \multicolumn{2}{c}{\begin{tabular}[c]{@{}c@{}}Combined\\(1NN)\cite{F52} \end{tabular} }  & \multicolumn{2}{c}{\begin{tabular}[c]{@{}c@{}}OS-\\CNN\cite{F25} \end{tabular}}            & \begin{tabular}[c]{@{}c@{}}Best:\\fcn-lstm\cite{F27} \end{tabular}    & \begin{tabular}[c]{@{}c@{}}Vanilla:RN-\\ Transformer \\ \cite{F26,F28}\end{tabular} & \begin{tabular}[c]{@{}c@{}}ResNet-\\ Transformer1 \\ \cite{F26}\end{tabular} & \begin{tabular}[c]{@{}c@{}}ResNet-\\ Transformer2 \\ \cite{F26}\end{tabular} & \begin{tabular}[c]{@{}c@{}}ResNet-\\ Transformer3 \\ \cite{F26}\end{tabular} & Ours              \\ \hline
Adiac                          & \multicolumn{2}{c}{0.857}           & \multicolumn{2}{c}{0.76}       & 0.841432          & \multicolumn{2}{c}{0.645}          & \multicolumn{2}{c}{0.838875}          & \textbf{0.869565} & 0.84399                & 0.849105              & 0.849105              & 0.849105              & 0.792839          \\
ArrowHead                      & \multicolumn{2}{c}{0.88}            & \multicolumn{2}{c}{0.817}      & 0.845714          & \multicolumn{2}{c}{0.817}          & \multicolumn{2}{c}{0.84}              & \textbf{0.925714} & 0.891429               & 0.891429              & 0.891429              & 0.891429              & 0.851429          \\
Beef                           & \multicolumn{2}{c}{\textbf{0.9}}    & \multicolumn{2}{c}{0.667}      & 0.7               & \multicolumn{2}{c}{0.6}            & \multicolumn{2}{c}{0.83333}           & \textbf{0.9}      & 0.866667               & 0.866667              & 0.866667             & 0.866667             & \textbf{0.9}      \\
BeetleFly                      & \multicolumn{2}{c}{0.95}            & \multicolumn{2}{c}{0.8}        & 0.8               & \multicolumn{2}{c}{0.8}            & \multicolumn{2}{c}{0.8}               & \textbf{1}        & \textbf{1}             & 0.95                  & 0.95                  & \textbf{1}            & \textbf{1}        \\
BirdChicken                    & \multicolumn{2}{c}{0.95}            & \multicolumn{2}{c}{0.9}        & 0.95              & \multicolumn{2}{c}{0.75}           & \multicolumn{2}{c}{0.9}               & 0.95              & \textbf{1}             & 0.9                   & \textbf{1}            & 0.7                   & \textbf{1}        \\
Car                            & \multicolumn{2}{c}{0.933}           & \multicolumn{2}{c}{0.85}       & 0.883333           & \multicolumn{2}{c}{0.8}            & \multicolumn{2}{c}{0.933333}          & \textbf{0.966667}          & 0.95                   & 0.883333              & 0.866667              & 0.3                   & 0.883333          \\
CBF                            & \multicolumn{2}{c}{\textbf{1}}      & \multicolumn{2}{c}{0.988}      & 0.998889          & \multicolumn{2}{c}{0.978}          & \multicolumn{2}{c}{0.998889}          & 0.996667          & \textbf{1}             & 0.997778              & \textbf{1}            & \textbf{1}            & \textbf{1}        \\
ChlorineConcentration          & \multicolumn{2}{c}{0.872}           & \multicolumn{2}{c}{0.688}      & 0.876563          & \multicolumn{2}{c}{0.588}          & \multicolumn{2}{c}{0.84974}           & 0.816146          & 0.849479               & 0.863281              & 0.409375              & 0.861719              & \textbf{0.894271} \\
CinCECGTorso                   & \multicolumn{2}{c}{\textbf{0.9949}} & \multicolumn{2}{c}{0.638}      & 0.853623          & \multicolumn{2}{c}{0.693}          & \multicolumn{2}{c}{0.830435}          & 0.904348          & 0.871739               & 0.656522              & 0.89058               & 0.31087               & 0.810145          \\
Coffee                         & \multicolumn{2}{c}{\textbf{1}}      & \multicolumn{2}{c}{\textbf{1}} & \textbf{1}        & \multicolumn{2}{c}{\textbf{1}}     & \multicolumn{2}{c}{\textbf{1}}        & \textbf{1}        & \textbf{1}             & \textbf{1}            & \textbf{1}            & \textbf{1}            & \textbf{1}        \\
CricketX                       & \multicolumn{2}{c}{0.821}           & \multicolumn{2}{c}{0.682}      & \textbf{0.853846} & \multicolumn{2}{c}{0.741}          & \multicolumn{2}{c}{0.846154}          & 0.792308          & 0.838462               & 0.8                   & 0.810256              & 0.8                   & 0.771795          \\
CricketY                       & \multicolumn{2}{c}{0.8256}          & \multicolumn{2}{c}{0.667}      & 0.851282          & \multicolumn{2}{c}{0.664}          & \multicolumn{2}{c}{\textbf{0.869231}} & 0.802564          & 0.838462               & 0.805128              & 0.825641              & 0.808766              & 0.789744          \\
CricketZ                       & \multicolumn{2}{c}{0.8154}          & \multicolumn{2}{c}{0.656}      & \textbf{0.861538} & \multicolumn{2}{c}{0.723}          & \multicolumn{2}{c}{\textbf{0.861538}} & 0.807692          & 0.820513               & 0.805128              & 0.128205              & 0.1                   & 0.787179          \\
DiatomSizeReduction            & \multicolumn{2}{c}{0.967}           & \multicolumn{2}{c}{0.974}      & 0.934641          & \multicolumn{2}{c}{0.967}          & \multicolumn{2}{c}{0.980392}          & 0.970588          & 0.993464               & \textbf{0.996732}     & 0.379085              & \textbf{0.996732}     & 0.980392          \\
DistalPhalanxOutlineAgeGroup   & \multicolumn{2}{c}{\textbf{0.835}}  & \multicolumn{2}{c}{0.727}      & 0.733813          & \multicolumn{2}{c}{0.669}          & \multicolumn{2}{c}{0.755396}          & 0.791367          & 0.81295                & 0.776978              & 0.467626              & 0.776978              & 0.719425          \\
DistalPhalanxOutlineCorrect    & \multicolumn{2}{c}{0.82}            & \multicolumn{2}{c}{0.764}      & 0.782608          & \multicolumn{2}{c}{0.683}          & \multicolumn{2}{c}{0.771739}          & 0.791367          & \textbf{0.822464}      & \textbf{0.822464}     & \textbf{0.822464}     & 0.793478              & 0.771739          \\
Earthquakes                    & \multicolumn{2}{c}{0.801}  & \multicolumn{2}{c}{0.748}      & 0.741007          & \multicolumn{2}{c}{0.64}           & \multicolumn{2}{c}{0.683453}          & \textbf{0.81295}           & 0.755396               & 0.755396              & 0.76259               & 0.755396              & 0.776978          \\
ECG200                         & \multicolumn{2}{c}{0.92}            & \multicolumn{2}{c}{0.83}       & 0.93              & \multicolumn{2}{c}{0.85}           & \multicolumn{2}{c}{0.91}              & 0.91              & 0.94                   & \textbf{0.95}         & 0.94                  & 0.93                  & 0.92              \\
ECG5000                        & \multicolumn{2}{c}{0.9482}          & \multicolumn{2}{c}{0.94}       & 0.940889          & \multicolumn{2}{c}{0.925}          & \multicolumn{2}{c}{0.940222}          & \textbf{0.948222} & 0.941556               & 0.943556              & 0.944222              & 0.940444              & 0.944444          \\
ECGFiveDays                    & \multicolumn{2}{c}{\textbf{1}}      & \multicolumn{2}{c}{\textbf{1}} & \textbf{1}        & \multicolumn{2}{c}{0.999}          & \multicolumn{2}{c}{\textbf{1}}        & 0.987224          & \textbf{1}             & \textbf{1}            & \textbf{1}            & \textbf{1}            & \textbf{1}        \\
FaceFour                       & \multicolumn{2}{c}{\textbf{1}}      & \multicolumn{2}{c}{0.83}       & 0.954545          & \multicolumn{2}{c}{0.864}          & \multicolumn{2}{c}{0.943182}          & 0.943182          & 0.954545               & 0.965909              & 0.977273              & 0.215909              & 0.924045          \\
FacesUCR                       & \multicolumn{2}{c}{0.958}           & \multicolumn{2}{c}{0.835}      & \textbf{0.97122}           & \multicolumn{2}{c}{0.86}           & \multicolumn{2}{c}{0.96439} & 0.941463          & 0.957561               & 0.947805              & 0.926829              & 0.95122               & 0.95122           \\
FordA                          & \multicolumn{2}{c}{0.9727} & \multicolumn{2}{c}{0.927}      & 0.961364          & \multicolumn{2}{c}{0.863}          & \multicolumn{2}{c}{0.958333}          & \textbf{0.976515}          & 0.948485               & 0.946212              & 0.517424              & 0.940909              & 0.939394          \\
FordB                          & \multicolumn{2}{c}{\textbf{0.9173}} & \multicolumn{2}{c}{0.798}      & 0.861782          & \multicolumn{2}{c}{0.748}          & \multicolumn{2}{c}{0.813580}          & 0.792593          & 0.838272               & 0.830864              & 0.838272              & 0.823457              & 0.823547          \\
GunPoint                       & \multicolumn{2}{c}{\textbf{1}}      & \multicolumn{2}{c}{0.987}      & \textbf{1}        & \multicolumn{2}{c}{0.833}          & \multicolumn{2}{c}{\textbf{1}}        & \textbf{1}        & \textbf{1}             & \textbf{1}            & \textbf{1}            & \textbf{1}            & \textbf{1}        \\
Ham                            & \multicolumn{2}{c}{0.781}           & \multicolumn{2}{c}{0.533}      & 0.714286          & \multicolumn{2}{c}{0.533}          & \multicolumn{2}{c}{0.714286}          & \textbf{0.809524} & 0.761905               & 0.780952              & 0.619048              & 0.514286              & \textbf{0.809524} \\
HandOutlines                   & \multicolumn{2}{c}{0.9487}          & \multicolumn{2}{c}{0.919}      & 0.954054          & \multicolumn{2}{c}{0.832}          & \multicolumn{2}{c}{\textbf{0.956757}} & 0.954054          & 0.937838               & 0.948649              & 0.835135              & 0.945946              & 0.894595          \\
Haptics                        & \multicolumn{2}{c}{0.551}           & \multicolumn{2}{c}{0.474}      & 0.548701          & \multicolumn{2}{c}{0.354}          & \multicolumn{2}{c}{0.512987}          & 0.558442          & 0.564935               & 0.545455              & \textbf{0.600649}     & 0.194805              & \textbf{0.600649} \\
Herring                        & \multicolumn{2}{c}{0.703}           & \multicolumn{2}{c}{0.578}      & 0.671875          & \multicolumn{2}{c}{0.563}          & \multicolumn{2}{c}{0.609375}          & \textbf{0.75}     & 0.703125               & 0.734375              & 0.65625               & 0.703125              & \textbf{0.75}     \\
InsectWingbeatSound            & \multicolumn{2}{c}{0.6525}          & \multicolumn{2}{c}{0.599}      & 0.638889          & \multicolumn{2}{c}{0.506}          & \multicolumn{2}{c}{0.637374}          & \textbf{0.668687} & 0.522222               & 0.642424              & 0.53859               & 0.536364              & 0.651515          \\
ItalyPowerDemand               & \multicolumn{2}{c}{0.97}            & \multicolumn{2}{c}{0.929}      & 0.965015          & \multicolumn{2}{c}{0.942}         & \multicolumn{2}{c}{0.947522}          & 0.963071          & 0.965015               & 0.969874              & 0.962099              & \textbf{0.971817}     & 0.964043          \\
Lightning2                     & \multicolumn{2}{c}{\textbf{0.8853}} & \multicolumn{2}{c}{0.787}      & 0.770492          & \multicolumn{2}{c}{0.885}          & \multicolumn{2}{c}{0.819672}          & 0.819672          & 0.852459               & 0.852459              & 0.754098              & 0.868852              & 0.836066          \\
Lightning7                     & \multicolumn{2}{c}{0.863}           & \multicolumn{2}{c}{0.74}       & 0.835616          & \multicolumn{2}{c}{0.795}          & \multicolumn{2}{c}{0.808219}          & 0.863014          & 0.821918               & 0.849315              & 0.383562              & 0.835616              & \textbf{0.90411}  \\
Mallat                         & \multicolumn{2}{c}{0.98}            & \multicolumn{2}{c}{0.916}      & 0.955224          & \multicolumn{2}{c}{\textbf{0.994}} & \multicolumn{2}{c}{0.963753}          & 0.98081           & 0.977399               & 0.975267              & 0.934328              & 0.979104              & 0.938593          \\
Meat                           & \multicolumn{2}{c}{\textbf{1}}      & \multicolumn{2}{c}{0.867}      & 0.933333          & \multicolumn{2}{c}{0.9}            & \multicolumn{2}{c}{0.983333}          & 0.883333          & \textbf{1}             & \textbf{1}            & \textbf{1}            & \textbf{1}            & \textbf{1}        \\
MedicalImages                  & \multicolumn{2}{c}{0.792}           & \multicolumn{2}{c}{0.725}      & 0.794737          & \multicolumn{2}{c}{0.693}          & \multicolumn{2}{c}{0.768421}          & \textbf{0.798684} & 0.780263               & 0.765789              & 0.759211              & 0.789474              & 0.793421          \\
MiddlePhalanxOutlineAgeGroup   & \multicolumn{2}{c}{\textbf{0.8144}} & \multicolumn{2}{c}{0.623}      & 0.551948          & \multicolumn{2}{c}{0.506}          & \multicolumn{2}{c}{0.538961}          & 0.668831          & 0.655844               & 0.662338              & 0.623377              & 0.662338              & 0.662388          \\
MiddlePhalanxOutlineCorrect    & \multicolumn{2}{c}{0.8076}          & \multicolumn{2}{c}{0.839}      & 0.817869          & \multicolumn{2}{c}{0.722}          & \multicolumn{2}{c}{0.807560}          & 0.841924          & \textbf{0.848797}      & \textbf{0.848797}     & \textbf{0.848797}     & 0.835052              & 0.744755          \\
MiddlePhalanxTW                & \multicolumn{2}{c}{0.612}           & \multicolumn{2}{c}{0.555}      & 0.512987          & \multicolumn{2}{c}{0.513}          & \multicolumn{2}{c}{0.564935}          & 0.603896          & 0.564935               & 0.577922              & 0.551948              & \textbf{0.623377}     & \textbf{0.62377}  \\
MoteStrain                     & \multicolumn{2}{c}{\textbf{0.95}}   & \multicolumn{2}{c}{0.823}      & 0.886581          & \multicolumn{2}{c}{0.853}          & \multicolumn{2}{c}{0.939297}          & 0.938498          & 0.940895               & 0.916933              & 0.9377                & 0.679712              & 0.875399          \\
OliveOil                       & \multicolumn{2}{c}{0.9333}          & \multicolumn{2}{c}{0.9}        & 0.833333          & \multicolumn{2}{c}{0.833}          & \multicolumn{2}{c}{0.833333}          & 0.766667          & \textbf{0.966667}      & 0.9                   & 0.933333              & 0.9                   & \textbf{0.966667} \\
Plane                          & \multicolumn{2}{c}{\textbf{1}}      & \multicolumn{2}{c}{0.981}      & \textbf{1}        & \multicolumn{2}{c}{\textbf{1}}     & \multicolumn{2}{c}{\textbf{1}}        & \textbf{1}        & \textbf{1}             & \textbf{1}            & \textbf{1}            & \textbf{1}            & \textbf{1}        \\
ProximalPhalanxOutlineAgeGroup & \multicolumn{2}{c}{0.8832}          & \multicolumn{2}{c}{0.839}      & 0.84878           & \multicolumn{2}{c}{0.805}          & \multicolumn{2}{c}{0.843902}          & 0.887805          & 0.887805               & \textbf{0.892683}     & 0.882927              & \textbf{0.892683}     & 0.878049          \\
ProximalPhalanxOutlineCorrect  & \multicolumn{2}{c}{0.918}           & \multicolumn{2}{c}{0.869}      & \textbf{0.931271}          & \multicolumn{2}{c}{0.801}          & \multicolumn{2}{c}{0.900344}          & \textbf{0.931271} & \textbf{0.931271}      & \textbf{0.931271}     & 0.683849              & 0.924399              & 0.910653          \\
ProximalPhalanxTW              & \multicolumn{2}{c}{0.815}           & \multicolumn{2}{c}{0.785}      & 0.77561           & \multicolumn{2}{c}{0.717}          & \multicolumn{2}{c}{0.775610}          & \textbf{0.843902} & 0.819512               & 0.814634              & 0.819512              & 0.819512              & 0.834146          \\
ShapeletSim                    & \multicolumn{2}{c}{\textbf{1}}      & \multicolumn{2}{c}{0.517}      & 0.955556          & \multicolumn{2}{c}{0.772}          & \multicolumn{2}{c}{0.827778}          & \textbf{1}        & \textbf{1}             & 0.91111               & 0.888889              & 0.9777778             & \textbf{1}        \\
ShapesAll                      & \multicolumn{2}{c}{0.9183}          & \multicolumn{2}{c}{0.837}      & 0.928333          & \multicolumn{2}{c}{0.823}          & \multicolumn{2}{c}{0.923333}          & 0.905             & 0.923333               & 0.876667              & 0.921667              & \textbf{0.933333}     & 0.876667          \\
SonyAIBORobotSurface1          & \multicolumn{2}{c}{0.985}           & \multicolumn{2}{c}{0.84}       & 0.8685552         & \multicolumn{2}{c}{0.825}          & \multicolumn{2}{c}{0.978369}          & 0.980525          & \textbf{0.988353}      & 0.978369              & 0.708819              & 0.985025              & 0.881864          \\
SonyAIBORobotSurface2          & \multicolumn{2}{c}{0.962}           & \multicolumn{2}{c}{0.832}      & 0.946485          & \multicolumn{2}{c}{0.885}          & \multicolumn{2}{c}{0.961175}          & 0.972718          & 0.976915               & 0.974816              & \textbf{0.98426}      & 0.976915              & 0.854145          \\
Strawberry                     & \multicolumn{2}{c}{0.976}           & \multicolumn{2}{c}{0.946}      & 0.983784          & \multicolumn{2}{c}{0.903}          & \multicolumn{2}{c}{0.981081}          & \textbf{0.986486} & \textbf{0.986486}      & \textbf{0.986486}     & \textbf{0.986486}     & \textbf{0.986486}     & \textbf{0.986486} \\
SwedishLeaf                    & \multicolumn{2}{c}{0.9664}          & \multicolumn{2}{c}{0.925}      & 0.9774            & \multicolumn{2}{c}{0.891}          & \multicolumn{2}{c}{0.9696}            & \textbf{0.9792}   & \textbf{0.9792}        & 0.9728                & 0.9696                & 0.9664                & 0.9376            \\
Symbols                        & \multicolumn{2}{c}{0.9668}          & \multicolumn{2}{c}{0.945}      & 0.980905          & \multicolumn{2}{c}{0.933}          & \multicolumn{2}{c}{0.976884}          & \textbf{0.98794}  & 0.9799                 & 0.970854              & 0.976884              & 0.252261              & 0.892462          \\
SyntheticControl               & \multicolumn{2}{c}{\textbf{1}}      & \multicolumn{2}{c}{0.977}      & 0.996667          & \multicolumn{2}{c}{0.977}          & \multicolumn{2}{c}{\textbf{1}}        & 0.993333          & \textbf{1}             & 0.996667              & \textbf{1}            & \textbf{1}            & \textbf{1}        \\
ToeSegmentation1               & \multicolumn{2}{c}{0.9737}          & \multicolumn{2}{c}{0.899}      & 0.964912          & \multicolumn{2}{c}{0.851}          & \multicolumn{2}{c}{0.956140}          & \textbf{0.991228} & 0.969298               & 0.969298              & 0.97807               & \textbf{0.991228}     & 0.982456          \\
ToeSegmentation2               & \multicolumn{2}{c}{0.9615}          & \multicolumn{2}{c}{0.9}        & 0.938462          & \multicolumn{2}{c}{0.9}            & \multicolumn{2}{c}{0.938462}          & 0.930769          & 0.976923               & 0.953846              & 0.953846              & \textbf{0.976923}     & 0.938462          \\
Trace                          & \multicolumn{2}{c}{\textbf{1}}      & \multicolumn{2}{c}{\textbf{1}} & \textbf{1}        & \multicolumn{2}{c}{\textbf{1}}     & \multicolumn{2}{c}{\textbf{1}}        & \textbf{1}        & \textbf{1}             & \textbf{1}            & \textbf{1}            & \textbf{1}            & \textbf{1}        \\

TwoPatterns                    & \multicolumn{2}{c}{\textbf{1}}      & \multicolumn{2}{c}{0.992}      & \textbf{1}        & \multicolumn{2}{c}{0.998}          & \multicolumn{2}{c}{\textbf{1}}        & 0.99675           & \textbf{1}             & \textbf{1}            & \textbf{1}            & \textbf{1}            & \textbf{1}        \\
TwoLeadECG                     & \multicolumn{2}{c}{\textbf{1}}      & \multicolumn{2}{c}{0.993}      & 0.99561           & \multicolumn{2}{c}{0.988}          & \multicolumn{2}{c}{0.999122}          & \textbf{1}        & \textbf{1}             & \textbf{1}            & \textbf{1}            & \textbf{1}            & \textbf{1}        \\
UWaveGestureLibraryAll         & \multicolumn{2}{c}{\textbf{0.9685}} & \multicolumn{2}{c}{0.865}      & 0.951982          & \multicolumn{2}{c}{0.838}          & \multicolumn{2}{c}{0.941653}          & 0.961195          & 0.856784               & 0.933277              & 0.939978              & 0.879118              & \textbf{0.9685}   \\
UWaveGestureLibraryX           & \multicolumn{2}{c}{0.8308}          & \multicolumn{2}{c}{0.784}      & 0.824958          & \multicolumn{2}{c}{0.762}          & \multicolumn{2}{c}{0.817700}          & \textbf{0.843663} & 0.780849               & 0.814629              & 0.810999              & 0.808766              & 0.815187          \\
UWaveGestureLibraryY           & \multicolumn{2}{c}{0.7585}          & \multicolumn{2}{c}{0.697}      & \textbf{0.767169}          & \multicolumn{2}{c}{0.666}          & \multicolumn{2}{c}{0.749860}          & 0.765215 & 0.664992               & 0.71636               & 0.671413              & 0.67895               & 0.752094          \\
UWaveGestureLibraryZ           & \multicolumn{2}{c}{0.7725}          & \multicolumn{2}{c}{0.729}      & 0.764098          & \multicolumn{2}{c}{0.679}          & \multicolumn{2}{c}{0.757956}          & \textbf{0.795924} & 0.756002               & 0.761027              & 0.760469              & 0.762144              & 0.757677          \\
Wafer                          & \multicolumn{2}{c}{\textbf{1}}      & \multicolumn{2}{c}{0.995}      & 0.99854           & \multicolumn{2}{c}{0.987}          & \multicolumn{2}{c}{0.998864}          & 0.998378          & 0.99854                & 0.998215              & 0.99854               & 0.999027              & \textbf{1}        \\
Wine                           & \multicolumn{2}{c}{0.889}           & \multicolumn{2}{c}{0.685}      & 0.611111          & \multicolumn{2}{c}{0.5}            & \multicolumn{2}{c}{0.555556}          & 0.833333          & 0.851852               & 0.87037               & 0.87037               & \textbf{0.907407}     & \textbf{0.907407} \\
WordSynonyms                   & \multicolumn{2}{c}{\textbf{0.779}}  & \multicolumn{2}{c}{0.641}      & 0.733542          & \multicolumn{2}{c}{0.633}          & \multicolumn{2}{c}{0.747649}          & 0.680251          & 0.661442               & 0.65047               & 0.636364              & 0.678683              & 0.659875          \\
ACSF1                          & \multicolumn{2}{c}{---}             & \multicolumn{2}{c}{0.73}       & 0.92              & \multicolumn{2}{c}{0.85}           & \multicolumn{2}{c}{0.92}              & 0.9               & \textbf{0.96}                   & 0.91                  & 0.93         & 0.17                  & 0.9               \\
BME                            & \multicolumn{2}{c}{---}             & \multicolumn{2}{c}{0.96}       & 0.99333           & \multicolumn{2}{c}{0.947}          & \multicolumn{2}{c}{\textbf{1}}        & 0.993333          & \textbf{1}             & \textbf{1}            & \textbf{1}            & \textbf{1}            & 0.986667          \\
Chinatown                      & \multicolumn{2}{c}{---}             & \multicolumn{2}{c}{0.962}      & 0.985423          & \multicolumn{2}{c}{0.936}          & \multicolumn{2}{c}{0.982609}          & 0.982609          & \textbf{0.985507}      & \textbf{0.985507}     & \textbf{0.985507}     & \textbf{0.985507}     & \textbf{0.985507} \\
Crop                           & \multicolumn{2}{c}{---}             & \multicolumn{2}{c}{0.727}      & 0.772202          & \multicolumn{2}{c}{0.695}          & \multicolumn{2}{c}{0.770179}          & 0.74494           & 0.743869               & 0.742738              & 0.746012              & 0.740476              & \textbf{0.774702} \\
DodgerLoopDay                  & \multicolumn{2}{c}{---}             & \multicolumn{2}{c}{---}        & 0.15              & \multicolumn{2}{c}{---}            & \multicolumn{2}{c}{0.5625}            & 0.6375            & 0.5357                 & 0.55                  & 0.4625                & 0.5                   & \textbf{0.675}    \\
DodgerLoopGame                 & \multicolumn{2}{c}{---}             & \multicolumn{2}{c}{---}        & 0.855072          & \multicolumn{2}{c}{---}            & \multicolumn{2}{c}{\textbf{0.920290}} & 0.898551          & 0.876812               & 0.891304              & 0.550725              & 0.905794              & 0.905797          \\
DodgerLoopWeekend              & \multicolumn{2}{c}{---}             & \multicolumn{2}{c}{---}        & 0.971014          & \multicolumn{2}{c}{---}            & \multicolumn{2}{c}{0.978261}          & 0.978261          & 0.963768               & 0.978261              & 0.949275              & 0.963768              & \textbf{0.985507} \\
GunPointAgeSpan                & \multicolumn{2}{c}{---}             & \multicolumn{2}{c}{0.987}      & 0.987342          & \multicolumn{2}{c}{0.991}          & \multicolumn{2}{c}{\textbf{1}}        & 0.996835          & 0.996835               & 0.996835              & \textbf{1}            & 0.848101              & 0.990506          \\
GunPointMaleVersusFemale       & \multicolumn{2}{c}{---}             & \multicolumn{2}{c}{\textbf{1}} & 0.993671          & \multicolumn{2}{c}{0.994}          & \multicolumn{2}{c}{\textbf{1}}        & \textbf{1}        & \textbf{1}             & \textbf{1}            & 0.996835              & 0.996835              & \textbf{1}        \\
GunPointOldVersusYoung         & \multicolumn{2}{c}{---}             & \multicolumn{2}{c}{\textbf{1}} & 0.965079          & \multicolumn{2}{c}{\textbf{1}}     & \multicolumn{2}{c}{\textbf{1}}        & 0.993651          & \textbf{1}             & \textbf{1}            & \textbf{1}            & \textbf{1}            & \textbf{1}        \\
InsectEPGRegularTrain          & \multicolumn{2}{c}{---}             & \multicolumn{2}{c}{\textbf{1}} & \textbf{1}        & \multicolumn{2}{c}{\textbf{1}}     & \multicolumn{2}{c}{\textbf{1}}        & 0.995984          & \textbf{1}             & \textbf{1}            & \textbf{1}            & \textbf{1}            & \textbf{1}        \\
InsectEPGSmallTrain            & \multicolumn{2}{c}{---}             & \multicolumn{2}{c}{\textbf{1}} & 0.943775          & \multicolumn{2}{c}{\textbf{1}}     & \multicolumn{2}{c}{0.473896}          & 0.935743          & 0.955823               & 0.927711              & 0.971888              & 0.477912              & \textbf{1}        \\
MelbournePedestrian            & \multicolumn{2}{c}{---}             & \multicolumn{2}{c}{0.947}      & 0.913899          & \multicolumn{2}{c}{0.914}          & \multicolumn{2}{c}{0.908163}          & 0.913061          & 0.912245               & 0.911837              & 0.904898              & 0.901633              & 0.95736           \\
PowerCons                      & \multicolumn{2}{c}{---}             & \multicolumn{2}{c}{0.933}      & 0.944444          & \multicolumn{2}{c}{0.894}          & \multicolumn{2}{c}{\textbf{1}}        & 0.994444          & 0.933333               & 0.9444444             & 0.927778              & 0.927778              & \textbf{1}        \\
Rock                           & \multicolumn{2}{c}{---}             & \multicolumn{2}{c}{0.54}       & 0.8               & \multicolumn{2}{c}{0.5}            & \multicolumn{2}{c}{0.56}              & \textbf{0.92}     & 0.78                   & \textbf{0.92}         & 0.82                  & 0.76                  & 0.88              \\
SemgHandGenderCh2              & \multicolumn{2}{c}{---}             & \multicolumn{2}{c}{0.84}       & 0.816667          & \multicolumn{2}{c}{0.863}          & \multicolumn{2}{c}{0.876667}          & 0.91              & 0.866667               & 0.916667              & 0.848333              & 0.651667              & \textbf{0.923333} \\
SemgHandMovementCh2            & \multicolumn{2}{c}{---}             & \multicolumn{2}{c}{0.516}      & 0.482222          & \multicolumn{2}{c}{0.709}          & \multicolumn{2}{c}{0.577778}          & 0.56              & 0.513333               & 0.504444              & 0.391111              & 0.468889              & \textbf{0.757778} \\
SemgHandSubjectCh2             & \multicolumn{2}{c}{---}             & \multicolumn{2}{c}{0.591}      & 0.824444          & \multicolumn{2}{c}{0.72}           & \multicolumn{2}{c}{0.713333}          & 0.873333          & 0.746667               & 0.74                  & 0.666667              & 0.788889              & \textbf{0.897778} \\
SmoothSubspace                 & \multicolumn{2}{c}{---}             & \multicolumn{2}{c}{0.94}       & 0.993333          & \multicolumn{2}{c}{0.833}          & \multicolumn{2}{c}{\textbf{1}}        & 0.98              & \textbf{1}             & \textbf{1}            & 0.99333               & \textbf{1}            & \textbf{1}        \\
UMD                            & \multicolumn{2}{c}{---}             & \multicolumn{2}{c}{0.986}      & 0.986111          & \multicolumn{2}{c}{0.958}          & \multicolumn{2}{c}{0.993056}          & 0.986111          & \textbf{1}             & \textbf{1}            & \textbf{1}            & \textbf{1}            & \textbf{1}       
\\\hline
Total &      \multicolumn{2}{c}{65}             & \multicolumn{2}{c}{82}      & 85          & \multicolumn{2}{c}{82}          & \multicolumn{2}{c}{85}          & 85          & 85             & 85             & 85       &85            &85  
\\
Win   &  \multicolumn{2}{c}{7}             &\multicolumn{2}{c}{ 0}       & 3             & \multicolumn{2}{c}{1}            & \multicolumn{2}{c}{2}      & \textbf{12}   & 2                      & 1                       & 1                       & 2                       & 8           \\
Tie   & \multicolumn{2}{c}{15}            &\multicolumn{2}{c}{ 7}       & 9             & \multicolumn{2}{c}{6}            & \multicolumn{2}{c}{16}     & 15            & 25                     & 22                      & 21                      & 23                      & \textbf{31} \\
Lose  & \multicolumn{2}{c}{43}            &\multicolumn{2}{c}{ 75}      & 73            & \multicolumn{2}{c}{75}           & \multicolumn{2}{c}{67}     & 58            & 58                     & 62                      & 63                      & 60                      & 46          \\
Best  & \multicolumn{2}{c}{22}            & \multicolumn{2}{c}{7}       & 12            &\multicolumn{2}{c}{ 7}            & \multicolumn{2}{c}{18}     & 27            & 27                     & 23                      & 22                      & 25                      & \textbf{39} 
\\
\bottomrule
   
\end{tabular}

}
\end{table}

\begin{table}[p]
  \caption{Results of eight traditional algorithms and our structure on 65 selected datasets.}
  \label{sample-table}
  \centering
  \Huge
  \resizebox{86mm}{97mm}{
\begin{tabular}{|c|ccccccccc|}
\hline
\multirow{2}{*}{Dataset}       & \multirow{2}{*}{$DD_{DTW}$ \cite{F71}} & \multirow{2}{*}{$DTD_{C}$ \cite{F71}} & \multirow{2}{*}{TSF \cite{F73}} & \multirow{2}{*}{TSBF \cite{F74}} & \multirow{2}{*}{LPS \cite{F61}} & \multirow{2}{*}{BOSS \cite{F51}} & \multirow{2}{*}{EE \cite{F16}} & \multirow{2}{*}{COTE \cite{F14}} & \multirow{2}{*}{Ours} \\
                               &                        &                       &                      &                       &                      &                       &                     &                       &                       \\ \hline
Adiac                          & 0.701                  & 0.701                 & 0.731                & 0.77                  & 0.77                 & 0.765                 & 0.665               & 0.79                  & \textbf{0.792839}     \\
ArrowHead                      & 0.789                  & 0.72                  & 0.726                & 0.754                 & 0.783                & 0.834                 & 0.811               & 0.811                 & \textbf{0.851429}     \\
Beef                           & 0.667                  & 0.667                 & 0.767                & 0.567                 & 0.6                  & 0.8                   & 0.633               & 0.867                 & \textbf{0.9}          \\
BeetleFly                      & 0.65                   & 0.65                  & 0.75                 & 0.8                   & 0.8                  & 0.9                   & 0.75                & 0.8                   & \textbf{1}            \\
BirdChicken                    & 0.85                   & 0.8                   & 0.8                  & 0.9                   & \textbf{1}           & 0.95                  & 0.8                 & 0.9                   & \textbf{1}            \\
Car                            & 0.8                    & 0.733                 & 0.767                & 0.783                 & 0.85                 & 0.833                 & 0.833               & \textbf{0.9}          & 0.883333              \\
CBF                            & 0.997                  & 0.98                  & 0.994                & 0.988                 & 0.999                & 0.998                 & 0.998               & 0.996                 & \textbf{1}            \\
ChlorineConcentration          & 0.708                  & 0.713                 & 0.72                 & 0.692                 & 0.608                & 0.661                 & 0.656               & 0.727                 & \textbf{0.894271}     \\
CinCECGTorso                   & 0.725                  & 0.725                 & 0.983                & 0.712                 & 0.736                & 0.887                 & 0.942               & \textbf{0.995}        & 0.810145              \\
Coffee                         & \textbf{1}             & \textbf{1}            & 0.964                & \textbf{1}            & \textbf{1}           & \textbf{1}            & \textbf{1}          & \textbf{1}            & \textbf{1}            \\
CricketX                       & 0.754                  & 0.754                 & 0.664                & 0.705                 & 0.697                & 0.736                 & 0.813               & \textbf{0.808}        & 0.771795              \\
CricketY                       & 0.777                  & 0.774                 & 0.672                & 0.736                 & 0.767                & 0.754                 & 0.805               & \textbf{0.826}        & 0.789744              \\
CricketZ                       & 0.774                  & 0.774                 & 0.672                & 0.715                 & 0.754                & 0.746                 & 0.782               & \textbf{0.815}        & 0.787179              \\
DiatomSizeReduction            & 0.967                  & 0.915                 & 0.931                & 0.899                 & 0.905                & 0.931                 & 0.944               & 0.928                 & \textbf{0.980392}     \\
DistalPhalanxOutlineAgeGroup   & 0.705                  & 0.662                 & \textbf{0.748}       & 0.712                 & 0.669                & \textbf{0.748}        & 0.691               & \textbf{0.748}        & 0.719425              \\
DistalPhalanxOutlineCorrect    & 0.732                  & 0.725                 & 0.772                & \textbf{0.783}        & 0.721                & 0.728                 & 0.728               & 0.761                 & 0.771739              \\
Earthquakes                    & 0.705                  & 0.705                 & 0.748                & 0.748                 & 0.64                 & 0.748                 & 0.741               & 0.748                 & \textbf{0.776978}     \\
ECG200                         & 0.83                   & 0.84                  & 0.87                 & 0.84                  & 0.86                 & 0.87                  & 0.88                & 0.88                  & \textbf{0.92}         \\
ECG5000                        & 0.924                  & 0.924                 & 0.939                & 0.94                  & 0.917                & 0.941                 & 0.939               & 0.941                 & \textbf{0.944444}     \\
ECGFiveDays                    & 0.769                  & 0.822                 & 0.956                & 0.877                 & 0.879                & \textbf{1}            & 0.82                & 0.999                 & \textbf{1}            \\
FaceFour                       & 0.83                   & 0.818                 & 0.932                & \textbf{1}            & 0.943                & \textbf{1}            & 0.909               & 0.898                 & 0.924045              \\
FacesUCR                       & 0.904                  & 0.908                 & 0.883                & 0.867                 & 0.926                & 0.957                 & 0.945               & 0.942                 & \textbf{0.95122}      \\
FordA                          & 0.723                  & 0.765                 & 0.815                & 0.85                  & 0.873                & 0.93                  & 0.736               & \textbf{0.957}        & 0.939394              \\
FordB                          & 0.667                  & 0.653                 & 0.688                & 0.599                 & 0.711                & 0.771                 & 0.662               & 0.804                 & \textbf{0.823547}     \\
GunPoint                       & 0.98                   & 0.987                 & 0.973                & 0.987                 & 0.993                & \textbf{1}            & 0.993               & \textbf{1}            & \textbf{1}            \\
Ham                            & 0.476                  & 0.552                 & 0.743                & 0.762                 & 0.562                & 0.667                 & 0.571               & 0.648                 & \textbf{0.809524}     \\
HandOutlines                   & 0.868                  & 0.865                 & \textbf{0.919}       & 0.854                 & 0.881                & 0.903                 & 0.889               & \textbf{0.919}        & 0.894595              \\
Haptics                        & 0.399                  & 0.399                 & 0.445                & 0.49                  & 0.432                & 0.461                 & 0.393               & 0.523                 & \textbf{0.600649}     \\
Herring                        & 0.547                  & 0.547                 & 0.609                & 0.641                 & 0.578                & 0.547                 & 0.578               & 0.625                 & \textbf{0.75}         \\
InsectWingbeatSound            & 0.355                  & 0.473                 & 0.633                & 0.625                 & 0.551                & 0.523                 & 0.595               & \textbf{0.653}        & 0.651515              \\
ItalyPowerDemand               & 0.95                   & 0.951                 & 0.96                 & 0.883                 & 0.923                & 0.909                 & 0.962               & 0.961                 & \textbf{0.964043}     \\
Lightning2                     & 0.869                  & 0.869                 & 0.803                & 0.738                 & 0.82                 & 0.836                 & \textbf{0.885}      & 0.869                 & 0.836066              \\
Lightning7                     & 0.671                  & 0.658                 & 0.753                & 0.726                 & 0.74                 & 0.685                 & 0.767               & 0.808                 & \textbf{0.90411}      \\
Mallat                         & 0.949                  & 0.927                 & 0.919                & \textbf{0.96}         & 0.908                & 0.938                 & 0.94                & 0.954                 & 0.938593              \\
Meat                           & 0.933                  & 0.933                 & 0.933                & 0.933                 & 0.883                & 0.9                   & 0.933               & 0.917                 & \textbf{1}            \\
MedicalImages                  & 0.737                  & 0.745                 & 0.755                & 0.705                 & 0.746                & 0.718                 & 0.742               & 0.758                 & \textbf{0.793421}     \\
MiddlePhalanxOutlineAgeGroup   & 0.539                  & 0.5                   & 0.578                & 0.578                 & 0.578                & 0.545                 & 0.558               & 0.636                 & \textbf{0.662388}     \\
MiddlePhalanxOutlineCorrect    & 0.732                  & 0.742                 & \textbf{0.828}       & 0.814                 & 0.773                & 0.78                  & 0.784               & 0.804                 & 0.744755              \\
MiddlePhalanxTW                & 0.487                  & 0.5                   & 0.565                & 0.597                 & 0.526                & 0.545                 & 0.513               & 0.571                 & \textbf{0.62377}      \\
MoteStrain                     & 0.833                  & 0.768                 & 0.869                & 0.903                 & 0.922                & 0.879                 & 0.883               & \textbf{0.937}        & 0.875399              \\
OliveOil                       & 0.833                  & 0.867                 & 0.867                & 0.833                 & 0.867                & 0.867                 & 0.867               & 0.9                   & \textbf{0.966667}     \\
Plane                          & \textbf{1}             & \textbf{1}            & \textbf{1}           & \textbf{1}            & \textbf{1}           & \textbf{1}            & \textbf{1}          & \textbf{1}            & \textbf{1}            \\
ProximalPhalanxOutlineAgeGroup & 0.8                    & 0.795                 & 0.834                & 0.849                 & 0.849                & 0.834                 & 0.805               & 0.854                 & \textbf{0.878049}     \\
ProximalPhalanxOutlineCorrect  & 0.794                  & 0.794                 & 0.849                & 0.828                 & 0.873                & 0.849                 & 0.808               & 0.869                 & \textbf{0.910653}     \\
ProximalPhalanxTW              & 0.766                  & 0.771                 & 0.8                  & 0.815                 & 0.81                 & 0.8                   & 0.766               & 0.78                  & \textbf{0.834146}     \\
ShapeletSim                    & 0.611                  & 0.6                   & \textbf{1}           & 0.478                 & 0.961                & \textbf{1}            & 0.817               & 0.961                 & \textbf{1}            \\
ShapesAll                      & 0.85                   & 0.838                 & \textbf{0.908}       & 0.792                 & 0.185                & \textbf{0.908}        & 0.867               & 0.892                 & 0.876667              \\
SonyAIBORobotSurface1          & 0.742                  & 0.71                  & 0.632                & 0.787                 & 0.795                & 0.632                 & 0.704               & 0.845                 & \textbf{0.881864}     \\
SonyAIBORobotSurface2          & 0.892                  & 0.892                 & 0.859                & 0.81                  & 0.778                & 0.859                 & 0.878               & \textbf{0.952}        & 0.854145              \\
Strawberry                     & 0.954                  & 0.957                 & 0.967                & 0.965                 & 0.952                & 0.976                 & 0.946               & 0.951                 & \textbf{0.986486}     \\
SwedishLeaf                    & 0.901                  & 0.896                 & 0.922                & 0.914                 & 0.915                & 0.922                 & 0.915               & \textbf{0.955}        & 0.9376                \\
Symbols                        & 0.953                  & 0.963                 & \textbf{0.967}       & 0.915                 & 0.946                & \textbf{0.967}        & 0.96                & 0.964                 & 0.892462              \\
SyntheticControl               & 0.993                  & 0.997                 & 0.987                & 0.993                 & 0.98                 & 0.967                 & 0.99                & \textbf{1}            & \textbf{1}            \\
ToeSegmentation1               & 0.807                  & 0.807                 & 0.741                & 0.781                 & 0.877                & 0.939                 & 0.829               & 0.974                 & \textbf{0.982456}     \\
ToeSegmentation2               & 0.746                  & 0.715                 & 0.815                & 0.8                   & 0.869                & 0.962                 & 0.892               & 0.915                 & \textbf{0.938462}     \\
Trace                          & \textbf{1}             & 0.99                  & 0.99                 & 0.98                  & 0.98                 & \textbf{1}            & 0.99                & \textbf{1}            & \textbf{1}            \\
TwoLeadECG                     & 0.978                  & 0.985                 & 0.759                & 0.866                 & 0.948                & 0.981                 & 0.971               & 0.993                 & \textbf{1}            \\
TwoPatterns                    & \textbf{1}             & \textbf{1}            & 0.991                & 0.976                 & 0.982                & 0.993                 & \textbf{1}          & \textbf{1}            & \textbf{1}            \\
UWaveGestureLibraryAll         & 0.935                  & 0.938                 & 0.957                & 0.926                 & 0.966                & 0.939                 & 0.968               & 0.964                 & \textbf{0.9685}       \\
UWaveGestureLibraryX           & 0.779                  & 0.775                 & 0.804                & 0.831                 & 0.829                & 0.762                 & 0.805               & 0.822                 & \textbf{0.815187}     \\
UWaveGestureLibraryY           & 0.716                  & 0.698                 & 0.727                & 0.736                 & 0.761                & 0.685                 & 0.726               & \textbf{0.759}        & 0.752094              \\
UWaveGestureLibraryZ           & 0.696                  & 0.679                 & 0.743                & \textbf{0.772}        & 0.768                & 0.695                 & 0.724               & 0.75                  & 0.757677              \\
Wafer                          & 0.98                   & 0.993                 & 0.996                & 0.995                 & 0.997                & 0.995                 & 0.997               & \textbf{1}            & \textbf{1}            \\
Wine                           & 0.574                  & 0.611                 & 0.63                 & 0.611                 & 0.63                 & 0.741                 & 0.574               & 0.648                 & \textbf{0.907407}     \\
WordSynonyms                   & 0.73                   & 0.73                  & 0.647                & 0.688                 & 0.701                & 0.638                 & \textbf{0.779}      & 0.757                 & 0.659875              \\ \hline
Total                          & 65                     & 65                    & 65                   & 65                    & 65                   & 65                    & 65                  & 65                    & 65                    \\
Win                            & 0                      & 0                     & 1                    & 3                     & 0                    & 0                     & 2                   & 11                    & \textbf{33}           \\
Tie                            & 4                      & 3                     & 6                    & 3                     & 3                    &\textbf{10}                   & 3                   & 9                     & \textbf{10}           \\
Lose                           & 61                     & 62                    & 58                   & 59                    & 62                   & 55                    & 60                  & 45                    & \textbf{22}           \\
Best                           & 4                      & 3                     & 7                    & 6                     & 3                    & 10                    & 5                   & 20                    & \textbf{43}           \\ \hline
\end{tabular}
}
\end{table}

\begin{table}[!htb]
  \caption{Results of different supervised algorithms on 20 'long' time series datasets.}
  \label{sample-table}
  \centering
  \Huge
  \resizebox{120mm}{32mm}{
\begin{tabular}{|c|cc|ccccccccccc|}
\toprule

Dataset           &Classes&SeriesLength  & \begin{tabular}[c]{@{}c@{}}Existing \\SOTA \end{tabular}  & \begin{tabular}[c]{@{}c@{}}USRL-\\FordA\cite{F52} \end{tabular} & \begin{tabular}[c]{@{}c@{}}Inception-\\Time\cite{F24} \end{tabular} & \begin{tabular}[c]{@{}c@{}}Combined\\(1NN)\cite{F52} \end{tabular} & \begin{tabular}[c]{@{}c@{}}OS-\\CNN\cite{F25} \end{tabular}  &  \begin{tabular}[c]{@{}c@{}}Best:\\ lstm-fcn \cite{F27} \end{tabular}     & \begin{tabular}[c]{@{}c@{}}Vanilla:RN-\\ Transformer \\\cite{F26,F28}\end{tabular}  & \begin{tabular}[c]{@{}c@{}}ResNet-\\ Transformer1 \\ \cite{F26}\end{tabular} & \begin{tabular}[c]{@{}c@{}}ResNet-\\ Transformer2 \\ \cite{F26}\end{tabular} & \begin{tabular}[c]{@{}c@{}}ResNet-\\ Transformer3 \\ \cite{F26}\end{tabular}  &  $Ours$                  \\\hline
BeetleFly        &2&512   & 0.95            & 0.8      & 0.8           & 0.8          & 0.8      & \textbf{1}        & \textbf{1}             & 0.95                    & 0.95                    & \textbf{1}                   & \textbf{1}        \\
BirdChicken     &2&512    & 0.95            & 0.9      & 0.95          & 0.75         & 0.9      & 0.95              & \textbf{1}             & 0.9                     & \textbf{1}              & 0.7                          & \textbf{1}        \\
CinCECGTorso &4&1639	&\textbf{0.9949}	&0.638&	0.853623&	0.693&	0.830435&	0.904348&	0.871739&	0.656522&	0.89058&	0.31087	&0.810145\\
Car             &4&577    & 0.933           & 0.85     & 0.88333       & 0.8          & 0.933333 & \textbf{0.966667} & 0.95                   & 0.883333                & 0.866667                & 0.3                       & 0.883333          \\
Earthquake      &2&512    & \textbf{0.801}  & 0.748    & 0.741007      & 0.64         & 0.683453 & 0.81295           & 0.755396               & 0.755396                & 0.76259                 & 0.755396                & 0.776978          \\
HandOutlines	&2&2709&0.9487	&0.919	&0.954054	&0.832	&\textbf{ 0.956757}	&0.954054	&0.937838	&0.948649	&0.835135	&0.945946	&0.894595\\
Haptics          &5&1092   & 0.551           & 0.474    & 0.548701      & 0.354        & 0.512987 & 0.558442          & 0.564935               & 0.545455                & \textbf{0.600649}       & 0.194805              & \textbf{0.600649} \\
Herring          &2&512   & 0.703           & 0.578    & 0.671875      & 0.563        & 0.609375 & \textbf{0.75}     & 0.703125               & 0.734375                & 0.65625                 & 0.703125                    & \textbf{0.75}     \\
Lighting2        &2&637   & \textbf{0.8853} & 0.787    & 0.770492      & 0.885        & 0.819672 & 0.819672          & 0.852459               & 0.852459                & 0.754098                & 0.868852                     & 0.836066          \\
Mallat	&8&1024&0.98&	0.916&	0.955224&	0.994&	0.963753&	\textbf{0.98081}&	0.977399&	0.975267&	0.934328&	0.979104&	0.938593\\
OliveOil         &4&570   & 0.9333          & 0.9      & 0.833333      & 0.833        & 0.833333 & 0.766667          & \textbf{0.966667}      & 0.9                     & 0.933333                & 0.9                     &  \textbf{0.966667} \\
ShapesAll	&60&512&0.9183	&0.837	&0.928333	&0.823	&0.923333	&0.905	&0.923333	&0.876667	&0.921667	&\textbf{0.933333}	&0.876667\\
UWaveGestureLibraryAll&8&945	&\textbf{0.9685}	&0.865	&0.951982	&0.838	&0.941653	&0.961195	&0.856784	&0.933277&	0.939978&	0.879118&	\textbf{0.9685}\\
ACSF1	&10&1460&---	&0.73	&0.92&	0.85&	0.92&	0.9&	\textbf{0.96}&	0.91&	0.93&	0.17&	0.9\\
InsectEPGRegularTrain&3&601	&---&\textbf{1}&	\textbf{1}&	\textbf{1}&	\textbf{1}&	0.995984&	\textbf{1}&\textbf{	1}&	\textbf{1}&	\textbf{1}&\textbf{1}\\
InsectEPGSmallTrain&3&601	&---&	\textbf{1}&	0.943775&	\textbf{1}&	0.473896&	0.935743&	0.955823&	0.927711&	0.971888&	0.477912	&\textbf{1}\\
SemgHandGendeCh2    &2&1500& ---             & 0.84     & 0.816667      & 0.863        & 0.876667 & 0.91              & 0.866667               & 0.916667                & 0.848333                & 0.651667               & \textbf{0.923333} \\
SemgHandMovementCh2 &6&1500& ---             & 0.516    & 0.482222      & 0.709        & 0.577778 & 0.56              & 0.513333               & 0.504444                & 0.391111                & 0.468889               & \textbf{0.757778} \\
SemgHandSubjectCh2 &5&1500 & ---             & 0.591    & 0.824444      & 0.72         & 0.713333 & 0.873333          & 0.746667               & 0.74                    & 0.666667                & 0.788889         & \textbf{0.897778} \\
Rock          &4&2844      & ---             & 0.54     & 0.8           & 0.5          & 0.56     & \textbf{0.92}     & 0.78                   & \textbf{0.92}                    & 0.82                    & 0.76                            & 0.88              \\\hline
 \multicolumn{3}{|c|}{Best}                 & 4               & 2        & 1             & 2            & 1        & 5                 & 5                      & 2                       & 3                       & 3                                     & \textbf{11}                 \\
 \multicolumn{3}{|c|}{MeanACC}              & ---             &0.77145	&0.8314531	&0.77235	&0.7914879	&0.87124325&	0.85910825&	0.8415111&	0.8336637&	0.6893953&	\textbf{0.8830541}        \\
\bottomrule 
\end{tabular}

}
\end{table}

\subsection{Evaluation of the RTFN-based Supervised Structure}
To evaluate the performance of the RTFN-based supervised structure, we compare it with a number of existing supervised algorithms against ‘win’/‘lose’/‘tie’, MeanACC, and AVG\_rank on 85 univariate and 30 multivariate datasets (see Tables 1-2), respectively. 

\subsubsection{Performance Comparison on Univariate Time Series}
Table 7 shows the top-1 accuracy results obtained by different supervised algorithms on the 85 selected datasets in the UCR2018 archive. For each dataset, the existing SOTA represents the best algorithm on that dataset \cite{F26}, including ConvTimeNet\cite{F23}, the elastic ensemble (EE) \cite{F16}, COTE \cite{F14}, BOSS \cite{F51} and so on; similarly, for each dataset, the Best:lstm-fcn is the best-performance approach on that dataset, e.g. it involves LSTM-FCN and ALSTM-FCN in \cite{F27}. Note that the existing SOTA did not consider the last 20 of the 85 datasets. 

It is seen that the proposed RTFN-based supervised structure performs the best in ‘tie’ and the second best in ‘win’, guaranteeing its first position in ‘best’. To be specific, ours wins in 8 cases and performs no worse than any other algorithm in 31 cases, which leads to 39 ‘best’ cases in the competition. Besides, the Best: lstm-fcn and the Vanilla: ResNet-Transformer achieve the second and third best performance, with respect to the ‘best’ score. Especially, the former is a winner of 12 datasets, indicating its outstanding performance. On the other hand, USRLFordA is the worst algorithm, with 7 ‘tie’ scores only. Besides, Table 8 compares eight traditional algorithms and our structure on 65 selected UCR datasets in terms of the top-1 accuracy. These traditional algorithms include $DD_{DTW}$ \cite{F71}, $DTD_{C}$ \cite{F71}, TSF \cite{F73}, TSBF \cite{F74}, LPS \cite{F61}, BOSS \cite{F51}, EE \cite{F16}, and COTE \cite{F14}. Our structure obtains better ‘win’, ‘tie’, ‘lose’, and ‘best’ scores than the rest of the algorithms.

In addition, to emphasize the performance of the proposed structure in long time series cases, we show the top-1 accuracy results of different algorithms in Table 9, where all the 20 ‘long’ time series datasets in Table 1 are tested. It is easily observed that the RTFN-based supervised structure, i.e., ours, is the best algorithm as it obtains the highest ‘Best’ and MeanACC values. That is because RTFN can mine sufficient local features and the relationships among them, thanks to the efficient cooperation of TFN and LSTMaN. Especially, LSTMaN relates different locations of the obtained representations and can thus capture their intrinsic regularizations during the data transmission process. 

On the other hand, Best:lstm-fcn also achieves decent performance regarding the ‘best’ value and mean accuracy because its LSTM helps to extract additional features from the input data to enrich the features obtained by the fcn networks. Vanilla:ResNet-Transformer is undoubtedly the one with the best performance among all the transformer-based networks for comparison. The reason is that the embedded attention mechanism can further link different positions of time series data and thus enhance the accuracy.

\subsubsection{Performance Comparison on Multivariate Time Series}

Table 10 shows the top-1 accuracy results obtained by different supervised algorithms on all 30 datasets in the UEA2018 archive. For each dataset, the existing SOTA represents the best algorithm on that dataset, including STC \cite{F53}, HC \cite{F54}, gRSF \cite{F55}, mv-ARF \cite{F60} and so on; similarly, Best:DTW, Best:DTWN, and Best:EDN are the best performance DTW-based (e.g., involving DTW$_{I}$ and DTW$_{A}$ in \cite{F18}), DTWN-based (e.g., involving DTW-1NN$_{I}$(n) and DTW-1NN$_{D}$(n) \cite{F18}.) and ED-NN-based (e.g., involving ED-1NN and ED-1NN (Normalized) \cite{F53}) approaches on that dataset, respectively. 

\begin{table}[h]
  \caption{Results of different supervised algorithms on all UEA2018 datasets. Abbreviations: MF – MLSTM-FCN \cite{F31}, WM – WEASEL + MUSE \cite{F56}.}
  \label{sample-table}
  \Huge
  \centering
  \resizebox{120mm}{50mm}{

\begin{tabular}{|c|cccccccccccccccc|}
\toprule
\begin{tabular}[c]{@{}l@{}}Dataset \\ Index \end{tabular}   & \begin{tabular}[c]{@{}c@{}}Existing \\SOTA \end{tabular}  & \begin{tabular}[c]{@{}c@{}}Best: \\ DTW \cite{F53}  \end{tabular}      &  \begin{tabular}[c]{@{}c@{}}Best: \\ DTWN \cite{F18}  \end{tabular}    & \begin{tabular}[c]{@{}c@{}}Best: \\ EDN \cite{F53}  \end{tabular} & LCEM  \cite{F18}         & XGBM \cite{F54}        & RFM \cite{F57}             & WM   \cite{F53}          & CBOSS \cite{F51}     & MLCN \cite{F31} & RISE  \cite{F59}    & TSF \cite{F73}           & TapNet  \cite{F58}       & MUSE \cite{F56}         & MF    \cite{F18,F31}         & Ours           \\ \hline
AWR      & 0.99                                                           & 0.987      & 0.987          & 0.97  &\textbf{ 0.993}          & 0.99       & 0.99           & \textbf{0.993} & 0.99       & 0.957 & 0.963      & 0.953          & 0.987          & \textbf{0.993} & 0.986          & \textbf{0.993} \\
AF       & 0.267                                                          & 0.267      & 0.267          & 0.267 & 0.467          & 0.4        & 0.333          & 0.267          & 0.267      & 0.333 & 0.267      & 0.2            & 0.333            & 0.4            & 0.2            & \textbf{0.533} \\
BM       & \textbf{1}                                                     & \textbf{1} & \textbf{1}     & 0.675 & \textbf{1}     & \textbf{1} & \textbf{1}     & \textbf{1}     & \textbf{1} & 0.875 & \textbf{1} & \textbf{1}     & \textbf{1}     & \textbf{1}     & \textbf{1}     & \textbf{1}     \\
CT       & 0.986                                                          & \textbf{1} & 0.969          & 0.964 & 0.979          & 0.983      & 0.985          & 0.99           & 0.986      & 0.917 & 0.986      & 0.931          &0.997     & 0.986         & 0.993          & 0.993          \\
CR       & ---                                                            & ---        & \textbf{1}     & 0.944 & 0.986          & 0.972      & 0.986          & 0.986          & ---        & ---   & ---        & ---            & 0.958            & --             & 0.986          & 0.986          \\
DDG      & 0.48                                                           & 0.58       & 0.6            & 0.275 & 0.375          & 0.4        & 0.4            & 0.575          & 0.48       & 0.38  & 0.22       & 0.46           & 0.575           & 0.58           & \textbf{0.675} & 0.6            \\
EW       & 0.749                                                          & 0.517      & 0.618          & 0.55  & 0.527          & 0.55       & \textbf{1}     & 0.89           & 0.511      & 0.33  & 0.626      & 0.712          &0.489             & --             & 0.809          & 0.685          \\
EP       & \textbf{1}                                                     & \textbf{1} & 0.978          & 0.667 & 0.986          & 0.978      & 0.986          & 0.993          & 0.979      & 0.732 & 0.979      & \textbf{1}     & 0.971          & 0.993          & 0.964          & 0.978          \\
EC       & \textbf{0.882}                                                 & 0.361      & 0.323          & 0.293 & 0.372          & 0.422      & 0.433          & 0.316          & 0.304      & 0.373 & 0.445      & 0.487          & 0.323          & 0.476          & 0.274          & 0.38           \\
ER       & 0.97                                                           & 0.926      & 0.133          & 0.133 & 0.2            & 0.133      & 0.133          & 0.133          & 0.919      & 0.941 & 0.881      & 0.859          & 0.133          & \textbf{0.974} & 0.133          & 0.941          \\
FD       & 0.656                                                          & 0.529      & 0.529          & 0.519 & 0.614          & 0.629     & 0.614          & 0.545          & 0.513      & 0.555 & 0.64       & 0.508          & 0.556          & 0.631          & 0.555          & \textbf{0.67}  \\
FM       & 0.582                                                          & 0.53       & 0.53           & 0.55  & 0.59           & 0.53       & 0.569          & 0.54           & 0.519      & 0.58  & 0.581      & 0.562          & 0.53          & 0.551          & 0.61           & \textbf{0.63}  \\
HMD      & 0.414                                                          & 0.224      & 0.306          & 0.279 & 0.649          & 0.541      & 0.5            & 0.378          & 0.292      & 0.544 & 0.481      & 0.312          & 0.378          & 0.362          & 0.378          & \textbf{0.662} \\
HW       & 0.478                                                          & 0.601      & \textbf{0.607} & 0.531 & 0.287          & 0.267      & 0.267          & 0.531          & 0.504      & 0.305 & 0.359      & 0.191          & 0.357          & 0.518          & 0.547          & 0.454          \\
HB       & 0.64                                                           & 0.604      & 0.717          & 0.62  & 0.761          & 0.693      & 0.8            & 0.727          & 0.564      & 0.458 & 0.535      & 0.518          & 0.751          & 0.515          & 0.714          & \textbf{0.785} \\
IW       & ---                                                            & ---        & 0.115          & 0.128 & 0.228          & 0.237      & 0.224          & ---            & ---        & ---   & ---        & ---            &0.208            & ---            & 0.105          & \textbf{0.467} \\
JV       & ---                                                            & ---        & 0.959          & 0.949 & 0.978 & 0.968      & 0.97           & 0.978 & ---        & ---   & ---        & ---            &0.965            & ---            &\textbf{0.992}          & 0.973          \\
LIB      & 0.9                                                            & 0.883      & 0.894          & 0.833 & 0.772          & 0.767      & 0.783          & 0.894          & 0.894      & 0.85  & 0.806      & 0.806          & 0.85          & 0.894          & \textbf{0.922} & \textbf{0.922} \\
LSST     & 0.391                                                          & 0.458      & 0.575          & 0.456 & \textbf{0.652}          & 0.633      & 0.612          & 0.628          & 0.458      & 0.39  & 0.161      & 0.265          & 0.568          & 0.435          & 0.646 & 0.451          \\
MI       & \textbf{0.61}                                                  & 0.59       & 0.51           & 0.51  & 0.6            & 0.46       & 0.55           & 0.5            & 0.39       & 0.51  & 0.48       & 0.55           &0.59            & ---            & 0.53           & 0.6            \\
NATO     & 0.889                                                          & 0.883      & 0.883          & 0.85  & 0.916          & 0.9        & 0.911          & 0.883          & 0.85       & 0.9  & 0.8        & 0.839          & 0.939          & 0.906          & 0.961          & \textbf{0.967} \\
PD       & 0.941                                                          & 0.977      & 0.977          & 0.939 & 0.977          & 0.951      & 0.951          & 0.969          & 0.939      & 0.979 & 0.892      & 0.831          & 0.98          & 0.967          & \textbf{0.987} & \textbf{0.987}           \\
PEMS     & 0.981                                                          & 0.981      & 0.734          & 0.705 & 0.942          & 0.983      & 0.983          & ---            & 0.73       & 0.745 & 0.982      & \textbf{0.994} & 0.751           & ---            & 0.653          & 0.936          \\
PS       & 0.321                                                          & 0.195      & 0.151          & 0.104 & 0.288          & 0.187      & 0.222          & 0.19           & 0.151      & 0.151 & 0.137      & 0.269          &0.175            & ---            & 0.275          & \textbf{0.33}  \\
RS       & 0.898                                                          & 0.891      & 0.868          & 0.842 &\textbf{ 0.941}          & 0.928      & 0.921 & 0.914          & 0.854      & 0.856 & 0.895      & 0.823          & 0.868          & 0.933          & 0.882          & 0.862          \\
SRS1     & 0.854                                                          & 0.806      & 0.775          & 0.771 & 0.839          & 0.829      & 0.826          & 0.744          & 0.765      & 0.908 & 0.84       & 0.724          &0.652 & 0.697          & 0.867          &\textbf{ 0.922}          \\
SRS2     & 0.533                                                          & 0.539      & 0.539          & 0.533 & 0.55           & 0.483      & 0.478          & 0.522          & 0.533      & 0.506 & 0.483      & 0.494          & 0.55          & 0.528          & 0.522          & \textbf{0.622} \\
SAD      & ---                                                            & ---        & 0.963          & 0.967 & 0.973          & 0.97       & 0.968          & 0.982          & ---        & ---   & ---        & ---            &0.983            & ---            & \textbf{0.994} & 0.986          \\
SWJ      & 0.467                                                          & 0.333      & 0.333          & 0.2   & 0.4            & 0.333      & 0.467          & 0.333          & 0.333      & 0.4   & 0.333      & 0.267          & 0.4          & 0.267          & 0.467          & \textbf{0.667} \\
UW       & 0.897                                                          & 0.903      & 0.903          & 0.881 & 0.897          & 0.894      & 0.9            & 0.903          & 0.869      & 0.859 & 0.775      & 0.684          & 0.894            & \textbf{0.931} & 0.857          & 0.903          \\
\hline
Best     & 4                                                              & 3          & 3              & 0     & 4              & 1          & 2              & 2              & 1          & 0     & 1          & 3              & 1              & 4              & 6              & \textbf{15}    \\
MeanACC  & 0.626                                                          & 0.586      & 0.658          & 0.597 & 0.691          & 0.668      & 0.692          & 0.643          & 0.553      & 0.532 & 0.552      & 0.541          & 0.657          & 0.502          & 0.683          & \textbf{0.763} \\
AVG\_rank &6.817 	&8.517 	&8.633 	&11.800 	&5.900 	&8.417 	&6.917 	&7.833 	&11.117 	&10.583 	&10.450 	&11.433 	&8.083 	&8.933 	&6.750 	&\textbf{3.817} \\
   \bottomrule
\end{tabular}
}
\end{table}

It is observed that the RTFN-based supervised structure, i.e., ours, performs the best among all algorithms for comparison since ours obtains the highest MeanACC and ‘best’ values, namely, 0.763 and 15, and the smallest AVG\_rank score, namely 3.817. Meanwhile, MF and LCEM are the second and third best algorithms, while Best:EDN is the worst one. The following explains why. On the one hand, ours takes advantage of TFN and LSTMaN to mine sufficient features from the input data. In particular, LSTMaN can discover the relationships among not only the representations associated with each variate, but also those associated with different variates. That is why ours achieves the best performance. Due to the efficient coordination of LSTM and FCN networks, MF (i.e., MLSTM-FCN) can learn significant connections among the features associated with multiple variates as many as possible. In the meantime, LCEM uses an explicit boosting-bagging approach to explore the interactions among the dimensions. It allows LCEM to capture enough complex relationships among dimensions at different timestamps. That is why MF and LCEM also achieve decent performance in supervised classification. On the other hand, Best:EDN is based on the traditional DTW-1NN approach, making it quite challenging to simultaneously focus on the useful representations in univariate data and their relationships in multivariate data. That is why deep learning approaches have attracted increasingly more research attention. The AVG\_rank results obtained by different supervised algorithms on 30 multivariate datasets are shown in Fig. 5.

\begin{figure}[h]
  \centering
\includegraphics[width=12cm]{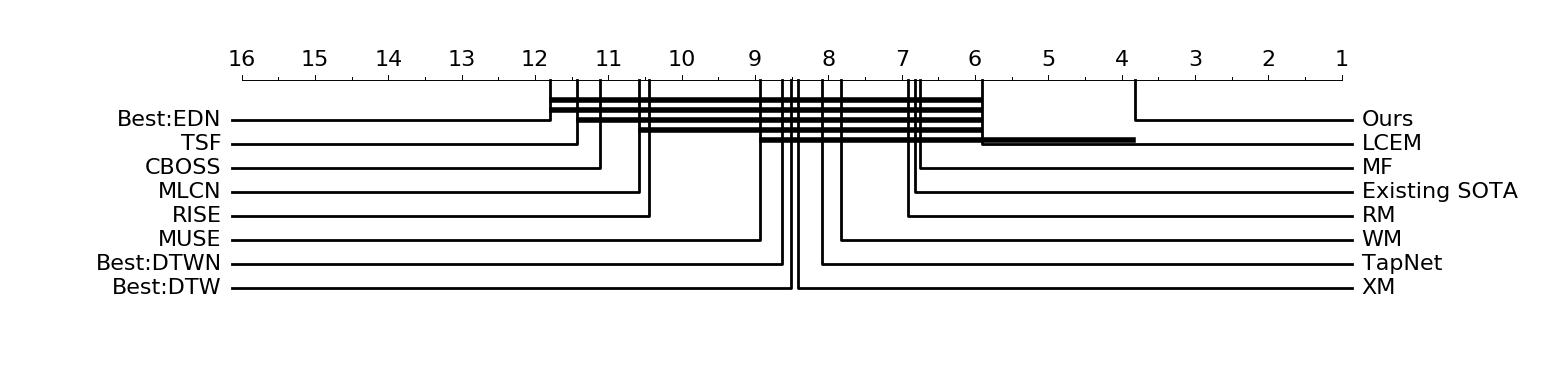}
\caption{Results of the AVG\_ranks of different algorithms on 30 multivariate datasets.}
\end{figure}

\subsection{Evaluation of the RTFN-based Unsupervised Clustering}
To evaluate the performance of the RTFN-based unsupervised clustering, we compare it with a number of state-of-the-art unsupervised algorithms against three performance metrics, namely ‘best’ based on the results of ‘win’/‘tie’/‘lose’, AVG RI, and AVG\_rank in Section 4.2.

\begin{table}[!htb]
  \caption{The RI results of different unsupervised algorithms on 36 selected datasets.}
  \label{Comparion}
  \centering
  \Huge
  \resizebox{120mm}{55mm}{
  \begin{tabular}{|c|cccccccccccccc|}
    \toprule

  Dataset&K-means\cite{F47}&UDFS\cite{F62}&NDFS\cite{F63}&RUFS\cite{F64}&RSFS\cite{F65}&KSC\cite{F66}&KDBA\cite{F67}&K-shape\cite{F68}&Ushapelet\cite{F69}&DTC\cite{F48}&DEC\cite{F70}&IDEC\cite{F49}&DTCR\cite{F15}&Ours
\\

    \midrule
   ArrowHead                          & 0.6095       & 0.7254    & 0.7381          & \textbf{0.7476} & 0.7108          & 0.7254          & 0.7222          & 0.7254          & 0.6460           & 0.6692    & 0.5817          & 0.6210          & 0.6868          & 0.6460          \\
Beef                           & 0.6713       & 0.6759    & 0.7034          & 0.7149          & 0.6975          & 0.7057          & 0.6713          & 0.5402          & 0.6966           & 0.6345    & 0.5954          & 0.6276          & \textbf{0.8046} & 0.7655          \\
BeetFly                        & 0.4789       & 0.4949    & 0.5579          & 0.6053          & 0.6516          & 0.6053          & 0.6052          & 0.6053          & 0.7314           & 0.5211    & 0.4947          & 0.6053          & \textbf{0.9000} & 0.6632          \\
BirdChicken                    & 0.4947       & 0.4947    & 0.7361          & 0.5579          & 0.6632          & 0.7316          & 0.6053          & 0.6632          & 0.5579           & 0.4947    & 0.4737          & 0.4789          & \textbf{0.8105} & 0.7316          \\
Car                            & 0.6345       & 0.6757    & 0.6260          & 0.6667          & 0.6708          & 0.6898          & 0.6254          & 0.7028          & 0.6418           & 0.6695    & 0.6859          & 0.6870          & \textbf{0.7501} & 0.7169          \\
ChlorineConcentration          & 0.5241       & 0.5282    & 0.5225          & 0.5330          & 0.5316          & 0.5256          & 0.5300          & 0.4111          & 0.5318           & 0.5353    & 0.5348          & 0.5350          & 0.5357          & \textbf{0.5367} \\
Coffee                         & 0.7460       & 0.8624    & \textbf{1.0000} & 0.5467          & \textbf{1.0000} & \textbf{1.0000} & 0.4851          & \textbf{1.0000} & \textbf{1.0000}  & 0.4841    & 0.4921          & 0.5767          & 0.9286          & 0.8624          \\
DiatomSizeReduction            & 0.9583       & 0.9583    & 0.9583          & 0.9333          & 0.9137          & \textbf{1.0000} & 0.9583          & \textbf{1.0000} & 0.7083           & 0.8792    & 0.9294          & 0.7347          & 0.9682          & 0.9583          \\
DistalPhalanxOutlineAgeGroup   & 0.6171       & 0.6531    & 0.6239          & 0.6252          & 0.6539          & 0.6535          & 0.6750          & 0.6020          & 0.6273           & 0.7812    & 0.7785          & 0.7786          & \textbf{0.7825} & 0.7786          \\
DistalPhalanxOutlineCorrect    & 0.5252       & 0.5362    & 0.5362          & 0.5252          & 0.5327          & 0.5235          & 0.5203          & 0.5252          & 0.5098           & 0.5010    & 0.5029          & 0.5330          & 0.6075          & \textbf{0.6095} \\
ECG200                         & 0.6315       & 0.6533    & 0.6315          & 0.7018          & 0.6916          & 0.6315          & 0.6018          & 0.7018          & 0.5758           & 0.6018    & 0.6422          & 0.6233          & 0.6648          & \textbf{0.7568} \\
ECGFIveDays                    & 0.4783       & 0.5020    & 0.5573          & 0.5020          & 0.5953          & 0.5257          & 0.5573          & 0.5020          & 0.5968           & 0.5016    & 0.5103          & 0.5114          & \textbf{0.9638} & 0.5964          \\
GunPoint                       & 0.4971       & 0.5029    & 0.5102          & \textbf{0.6498} & 0.4994          & 0.4971          & 0.5420          & 0.6278          & 0.6278           & 0.5400    & 0.4981          & 0.4974          & 0.6398          & 0.6471          \\
Ham                            & 0.5025       & 0.5219    & 0.5362          & 0.5107          & 0.5127          & 0.5362          & 0.5141          & 0.5311          & 0.5362           & 0.5648    & \textbf{0.5963} & 0.4956          & 0.5362          & \textbf{0.5963} \\
Herring                        & 0.4965       & 0.5099    & 0.5164          & 0.5238          & 0.5151          & 0.4940          & 0.5164          & 0.4965          & 0.5417           & 0.5045    & 0.5099          & 0.5099          & \textbf{0.5790} & 0.5322          \\
Lighting2                      & 0.4966       & 0.5119    & 0.5373          & 0.5729          & 0.5269          & 0.6263          & 0.5119          & 0.6548          & 0.5192           & 0.5770    & 0.5311          & 0.5519          & 0.5913          & \textbf{0.6230} \\
Meat                           & 0.6595       & 0.6483    & 0.6635          & 0.6578          & 0.6657          & 0.6723          & 0.6816          & 0.6575          & 0.6742           & 0.3220    & 0.6475          & 0.6220          & \textbf{0.9763} & 0.7175          \\
MiddlePhalanxOutlineAgeGroup   & 0.5351       & 0.5269    & 0.5350          & 0.5315          & 0.5473          & 0.5364          & 0.5513          & 0.5105          & 0.5396           & 0.5757    & 0.7059          & 0.6800          & \textbf{0.7982} & 0.6800          \\
MiddlePhalanxOutlineCorrect    & 0.5000       & 0.5431    & 0.5047          & 0.5114          & 0.5149          & 0.5014          & 0.5563          & 0.5114          & 0.5218           & 0.5272    & 0.5423          & 0.5423          & 0.5617          & \textbf{0.5909} \\
MiddlePhalanxTW                & 0.0983       & 0.1225    & 0.1919          & 0.7920          & 0.8062          & 0.8187          & 0.8046          & 0.6213          & 0.7920           & 0.7115    & 0.8590          & 0.8626          & \textbf{0.8638} & 0.8062          \\
MoteStrain                     & 0.4947       & 0.5579    & 0.6053          & 0.5579          & 0.6168          & 0.6632          & 0.4789          & 0.6053          & 0.4789           & 0.5062    & 0.7435          & 0.7342          & \textbf{0.7686} & 0.6826          \\
OSULeaf                        & 0.5615       & 0.5372    & 0.5622          & 0.5497          & 0.5665          & 0.5714          & 0.5541          & 0.5538          & 0.5525           & 0.7329    & 0.7484          & 0.7607          & \textbf{0.7739} & 0.7513          \\
Plane                          & 0.9081       & 0.8949    & 0.8954          & 0.9220          & 0.9314          & 0.9603          & 0.9225          & 0.9901          & \textbf{1.0000}  & 0.9040    & 0.9447          & 0.9447          & 0.9549          & 0.9619          \\
ProximalPhalanxOutlineAgeGroup & 0.5288       & 0.4997    & 0.5463          & 0.5780          & 0.5384          & 0.5305          & 0.5192          & 0.5617          & 0.5206           & 0.7430    & 0.4263          & 0.8091          & 0.8091          & \textbf{0.8180} \\
ProximalPhalanxTW              & 0.4789       & 0.4947    & 0.6053          & 0.5579          & 0.5211          & 0.6053          & 0.5211          & 0.5211          & 0.4789           & 0.8380    & 0.8189          & \textbf{0.9030} & 0.9023          & 0.8180          \\
SonyAIBORobotSurface1          & 0.7721       & 0.7695    & 0.7721          & 0.7787          & 0.7928          & 0.7726          & 0.7988          & 0.8084          & 0.7639           & 0.5563    & 0.5732          & 0.6900          & \textbf{0.8769} & 0.7812          \\
SonyAIBORobotSurface2          & 0.8697       & 0.8745    & 0.8865          & 0.8756          & 0.8948          & \textbf{0.9039} & 0.8684          & 0.5617          & 0.8770           & 0.7012    & 0.6514          & 0.6572          & 0.8354          & 0.7066          \\
SwedishLeaf                    & 0.4987       & 0.4923    & 0.5500          & 0.5192          & 0.5038          & 0.4923          & 0.5500          & 0.5533          & 0.6154           & 0.8871    & 0.8837          & 0.8893          & \textbf{0.9223} & 0.8893          \\
Symbols                        & 0.8810       & 0.8548    & 0.8562          & 0.8525          & 0.9060          & 0.8982          & \textbf{0.9774} & 0.8373          & 0.9603           & 0.9053    & 0.8841          & 0.8857          & 0.9168          & 0.9053          \\
ToeSegmentation1               & 0.4873       & 0.4921    & 0.5873          & 0.5429          & 0.4968          & 0.5000          & \textbf{0.6143} & \textbf{0.6143} & 0.5873           & 0.5077    & 0.4984          & 0.5017          & 0.5659          & \textbf{0.6143} \\
ToeSegmentation2               & 0.5257       & 0.5257    & 0.5968          & 0.5968          & 0.5826          & 0.5257          & 0.5573          & 0.5257          & 0.5020           & 0.5348    & 0.4991          & 0.4991          & \textbf{0.8286} & 0.7825          \\
TwoPatterns                    & 0.8529       & 0.8259    & 0.8530          & 0.8385          &\textbf {0.8588}          & 0.8585 & 0.8446          & 0.8046          & 0.7757           & 0.6251    & 0.6293          & 0.6338          & 0.6984          & 0.6298          \\
TwoLeadECG                     & 0.5476       & 0.5495    & 0.6328          & \textbf{0.8246} & 0.5635          & 0.5464          & 0.5476          & \textbf{0.8246} & 0.5404           & 0.5116    & 0.5007          & 0.5016          & 0.7114          & 0.6289          \\
Wafer                          & 0.4925       & 0.4925    & 0.5263          & 0.5263          & 0.4925          & 0.4925          & 0.4925          & 0.4925          & 0.4925           & 0.5324    & 0.5679          & 0.5597          & 0.7338          & \textbf{0.8093} \\
Wine                           & 0.4984       & 0.4987    & 0.5123          & 0.5021          & 0.5033          & 0.5006          & 0.5064          & 0.5001          & 0.5033           & 0.4906    & 0.4913          & 0.5157          & \textbf{0.6271} & 0.5471          \\
WordSyonyms                    & 0.8775       & 0.8697    & 0.8760          & 0.8861          & 0.8817          & 0.8727          & 0.8159          & 0.7844          & 0.8230           & 0.8855    & 0.8893          & 0.8947          & \textbf{0.8984} & 0.8973          \\\hline
 Best                           & 0            & 0         & 1               & 3               & 2               & 3               & 2               & 4               & 2                & 0         & 1               & 1               & \textbf{17}     & 9               \\
AVG RI                        & 0.5957       & 0.6077    & 0.6403          & 0.6477          & 0.6542          & 0.6582          & 0.6335          & 0.6419          & 0.6402           & 0.6238    & 0.6351          & 0.6515          & \textbf{0.7714} & 0.7233          \\
AVG\_rank                     & 10.8194      & 9.6944    & 7.2361          & 7.4306          & 6.8750          & 7.1667          & 7.9722          & 8.0000          & 8.1250           & 8.6111    & 8.7778          & 7.6389          & \textbf{2.7778} & 3.6250         
\\

    \bottomrule
  \end{tabular}
  }
\end{table}

Following some well-recognized research works \cite{F49,F15,F48,F68,F66}, we select 36 representative datasets from the UCR2018 archive for performance evaluation, and they are marked with 'YES' in Table 1. The RI results obtained by different unsupervised algorithms on the 36 datasets are shown in Table 11.

DTCR and our RTFN-based unsupervised clustering rank the best and second best among all algorithms for comparison. DTCR takes advantage of a seq2seq structure to explore sufficient temporal features for a K-means classifier. This algorithm uses the classifier's loss to update its model parameters, encouraging the representations extracted from data to form a cluster structure. That is why DTCR is good at mining cluster-specific representations from input data. Its complicated structure is, however, for addressing cluster-specific problems only. On the contrary, ours simply adopts an auto-encoding structure to update our model's parameters and utilizes a K-means algorithm to classify the features obtained by RTFN. Although it is quite simple in structure, ours achieves decent performance on the 36 selected datasets, which depends on the strong feature extraction ability of RTFN.

In order to further evaluate the effectiveness of RTFN in unsupervised clustering, we compare the RTFN-based K-means algorithm with a separate K-means algorithm on the 36 datasets above and show the RI results in Fig. 6. One can see that ours outperforms the separate K-means algorithm on all but two datasets. That is because our unsupervised clustering is provided with sufficient features obtained by the proposed RTFN, especially those hiding deeply in the input data beyond the exploration abilities of ordinary feature extraction networks. The AVG\_rank results of all unsupervised algorithms for comparison are shown in Fig. 7.

\begin{figure}[h]
  \centering
\includegraphics[width=12cm]{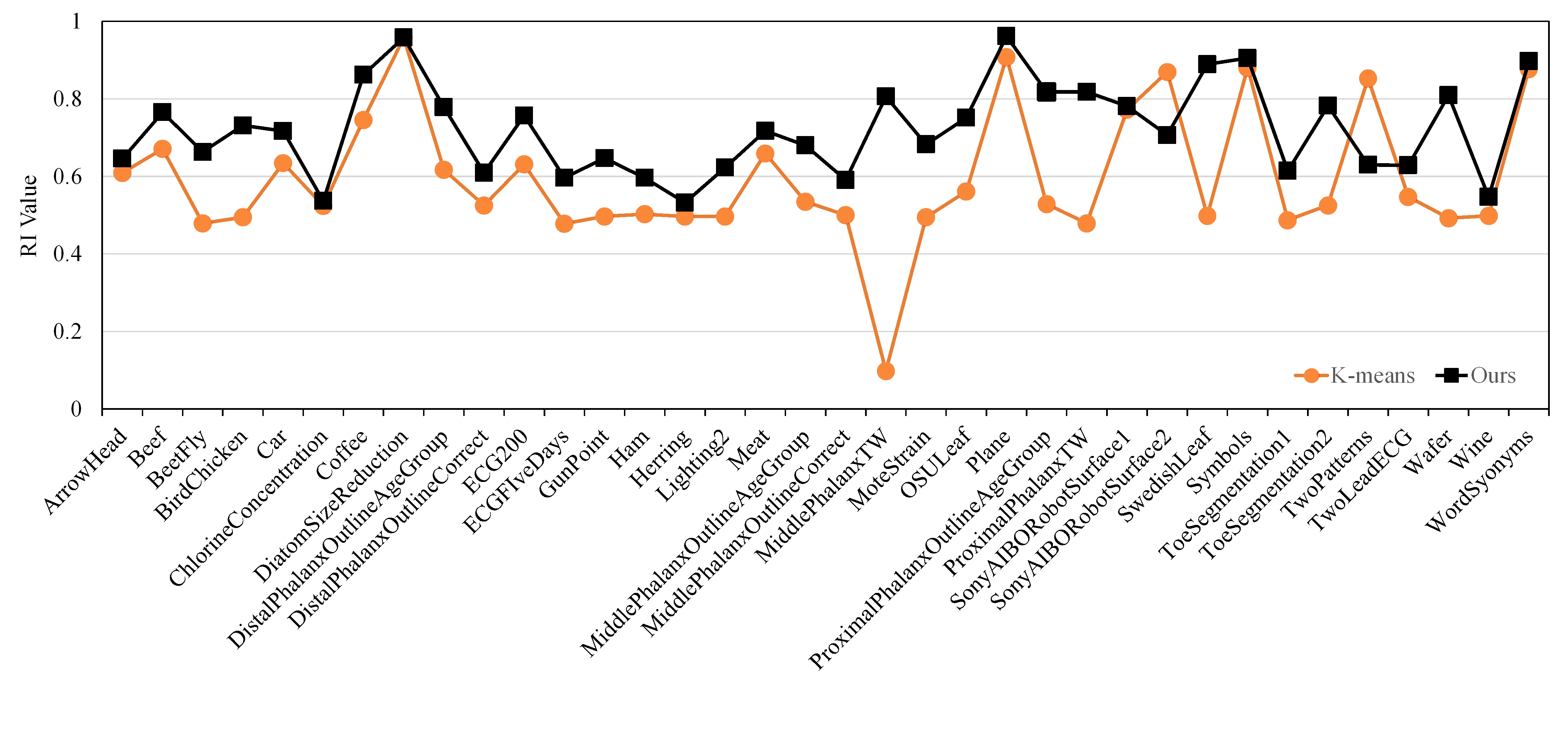}
\caption{The RI values obtained by the K-means algorithm and ours.}
\end{figure}

\begin{figure}[h]
  \centering
\includegraphics[height=2.5cm,width=12cm]{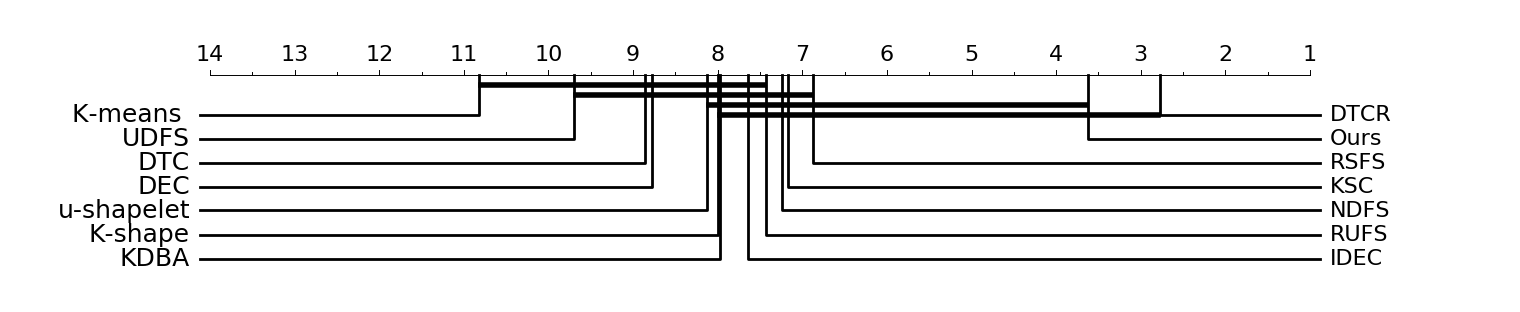}
\caption{AVG\_ranks of different unsupervised algorithms.}
\end{figure}

\section{Conclusion}

In the proposed RTFN, the temporal feature network is responsible for extracting local features, while the LSTM-based attention network aims at discovering intrinsic relationships among the representations learned from data. Experimental results demonstrate that our RTFN achieves decent performance in both supervised classification and unsupervised clustering. Specifically, our RTFN-based supervised algorithm performs the best on 39 out of 85 univariate datasets in the UCR2018 archive and 15 out of 30 multivariate datasets in the UEA2018 archive, respectively, compared with the latest results from the supervised classification community. In particular, ours wins 11 out of 20 long univariate dataset cases. Our RTFN-based unsupervised algorithm is the second best when considering all 36 datasets. Last but not least, the experimental results also indicate that RTFN has good potential to be embedded in other learning frameworks to handle time series problems of various domains in the real world.

\section{Acknowledgements}
This work was supported in part by National Natural Science Foundation of China (No. 61802319, No. 62002300), China Postdoctoral Science Foundation (2019M663552), the Fundamental Research Funds for the Central Universities, and China Scholarship Council, P. R. China.

\normalem
\bibliography{mybibfile}

\begin{thebibliography}{10}

\bibitem{F77}
J.~Audibert, P.~Michiardi, F.~Guyard, S.~Marti, and M.~A. Zuluaga.
\newblock Usad: unsupervised anomaly detection on multivariate time series.
\newblock {\em in Proc. ACM KDD'20}, pages 23--27, 2020.

\bibitem{F4}
A.~Bagnall, H.~A. Dau, J.~Lines, M.~Flynn, J.~Large, A.~Bostrom, P.~Southam,
  and E.~Keogh.
\newblock The uea multivariate time series classification archive, 2018.
\newblock {\em arXiv preprint arXiv: 1811.00075}, 2018.

\bibitem{F12}
A.~Bagnall., J.~Lines, A.~Bostrom, J.~Large, and E.~Keogh.
\newblock The great time series classification bake off: a review and
  experimental evaluation of recent algorithmic advances.
\newblock {\em Data Min. Knowl. Disc.}, 31:606--660, 2017.

\bibitem{F76}
M.~G. Baydogan and G.~Runger.
\newblock Learning a symbolic representation for multivariate time series
  classification.
\newblock {\em Data Min. Knowl. Disc.}, 29:400–422, 2015.

\bibitem{F61}
M.~G. Baydogan and G.~Runger.
\newblock Time series representation and similarity based on local
  autopatterns.
\newblock {\em Data Min. Knowl. Disc.}, 30:476--509, 2016.

\bibitem{F74}
M.~G. Baydogan, G.~Runger, and E.~Tuv.
\newblock A bag-of-features framework to classify time series.
\newblock {\em IEEE Trans. Pattern Anal.}, 35(11):2796–2802, 2013.

\bibitem{F6}
Z.~Che, S.~Purushotham, K.~Cho, D.~Sontag, and Y.~Liu.
\newblock Recurrent neural networks for multivariate time series with missing
  values.
\newblock {\em Sci. Rep.}, 8:6085, 2018.

\bibitem{F30}
J.~Chen, Z.~Xiao, H.~Xing, P.~Dai, S.~Luo, and M.~A. Iqbal.
\newblock Stdpg: a spatio-temporal deterministic policy gradient agent for
  dynamic routing in sdn.
\newblock {\em in Proc. IEEE ICC 2020}, pages 1--6, 2020.

\bibitem{F44}
X.~{Chen}, J.~{Yu}, and Z.~{Wu}.
\newblock Temporally identity-aware ssd with attentional lstm.
\newblock {\em IEEE Trans. Cybernetics}, 50(6):2674--2686, 2020.

\bibitem{F7}
H.~A. Dau, A.~Bagnall, C.-C.~M. Yeh, Y.~Zhu, S.~Gharghabi, C.~A.
  Ratanamahatana, and E.~Keogh.
\newblock The ucr time series archive.
\newblock {\em arXiv preprint arXiv: 1810.07758}, 2018.

\bibitem{F73}
H.~Deng, G.~Runger, E.~Tuv, and M.~Vladimir.
\newblock A time series forest for classification and feature extraction.
\newblock {\em Inform. Sciences}, 239:142--153, 2013.

\bibitem{F9}
J.~Deng, A.~Berg, S.~Satheesh, H.~Su, A.~Khosla, and L.~Fei-Fei.
\newblock Imagenet dataset.
\newblock {\em in Proc. ILSVRC-2012}, 2012.
\newblock URL http://www.image-net.org/challenges/LSVRC/2012/.

\bibitem{F18}
K.~Fauvel, {\'E}.~Fromont, V.~Masson, P.~Faverdin, and A.~Termier.
\newblock Local cascade ensemble for multivariate data classification.
\newblock {\em arXiv preprint arXiv:2005.03645}, 2020.

\bibitem{F36}
H.~I. Fawaz, G.~Forestier, J.~Weber, L.~Idoumghar, and P.-A. Muller.
\newblock Adversarial attack on deep neural networks for time series
  classification.
\newblock {\em in Proc. IJCNN 2019}, pages 1--8, 2019.

\bibitem{F14}
H.~I. Fawaz, G.~Forestier, J.~Weber, L.~Idoumghar, and P.~A. Muller.
\newblock Deep learning for time series classification: a reviewer.
\newblock {\em Data Min. Knowl. Disc.}, 33:917--963, 2019.

\bibitem{F24}
H.~I. Fawaz, B.~Lucas, G.~Forestier, C.~Pelletier, D.~F. Schmidt, J.~Weber,
  G.I. Webb, L.~Idoumghar, P.-A. Muller, and F.~Petitjean.
\newblock Inceptiontime: finding alexnet for time series classification.
\newblock {\em Data Min. Knowl. Disc.}, 34:1936--1962, 2020.

\bibitem{F52}
J.-Y. Franceschi, A.~Dieuleveut, and M.~Jaggi.
\newblock Unsupervised scalable representation learning for multivariate time
  series.
\newblock {\em in Proc. NeurIPS 2019}, pages 1--25, 2019.

\bibitem{F40}
Z.~Geng, G.~Chen, Y.~Han, G.~Lu, and F.~Li.
\newblock Semantic relation extraction using sequential and tree-structured
  lstm with attention.
\newblock {\em Inform. Sciences}, 509:183--192, 2020.

\bibitem{F49}
X.~Guo, L.~Gao, X.~Liu, and J.~Yin.
\newblock Improved deep embedded clustering with local structure preservation.
\newblock {\em in Proc. IJCAI 2017}, pages 1753--1759, 2017.

\bibitem{F29}
J.~Han, H.~Liu, M.~Wang, Z.~Li, and Y.~Zhang.
\newblock Era-lstm: an efficient reram-based architecture for long short-term
  memory.
\newblock {\em IEEE Trans. Parallel Distrib. Syst.}, 31(6):1328--1342, 2020.

\bibitem{F47}
J.~A. Hartigan and M.~A. Wong.
\newblock Algorithm as 136: a k-means clustering algorithm.
\newblock {\em J. Roy. Statist. Soc. Ser. C.}, 1(28):100--108, 1979.

\bibitem{F72}
S.~Hochreiter and J.~Schmidhuber.
\newblock Long short-term memory.
\newblock {\em Neural Comput.}, 9(8):1735--1780, 1997.

\bibitem{F26}
S.~H. Huang, L.~Xu, and C.~Jiang.
\newblock Residual attention net for superior cross-domain time sequence
  modeling.
\newblock {\em arXiv preprint arXiv: 2001.04077}, 2020.

\bibitem{F27}
F.~Karim, S.~Majumdar, and H.~Darabi.
\newblock Insights into lstm fully convolutional networks for time series
  classification.
\newblock {\em IEEE Access}, 7:1328--1342, 2019.

\bibitem{F31}
F.~Karim, S.~Majumdar, H.~Darabi, and S.~Harford.
\newblock Multivariate lstm-fcns for time series classification.
\newblock {\em Neural Networks}, 116:237--245, 2019.

\bibitem{F55}
I.~Karlsson, P.~Papapetrou, and H.~Bostrôm.
\newblock Generalized random shapelet forests.
\newblock {\em Data Min. Knowl. Disc.}, 30(5):1053--1083, 2016.

\bibitem{F23}
K.~Kashiparekh, J.~Narwariya, P.~Malhotra, L.~Vig, and G.~Shroff.
\newblock Convtimenet: A pre-trained deep convolutional neural network for time
  series classification.
\newblock {\em in Proc. IJCNN 2019}, pages 1--8, 2019.

\bibitem{F13}
M.~L{\"a}ngkvist, L.~Karlsson, and A.~Loutf.
\newblock A review of unsupervised feature learning and deep learning for
  time-series modeling.
\newblock {\em Pattern Recogn. Lett.}, 42(1):11--24, 2014.

\bibitem{F51}
J.~Large, A.~Bagnall, S.~Malinowski, and R.~Tavenard.
\newblock From bop to boss and beyond: time series classification with
  dictionary based classifier.
\newblock {\em arXiv preprint arXiv:1809.06751}, 2018.

\bibitem{F54}
J.~Large, J.~Lines, and A.~Bagnall.
\newblock A probabilistic classifier ensemble weighting scheme based on cross
  validated accuracy estimates.
\newblock {\em Data Min. Knowl. Disc.}, 33:1674--1709, 2019.

\bibitem{F22}
Y.~LeCun, Y.~Bengio, and G.~Hinton.
\newblock Deep learning.
\newblock {\em Nature}, pages 436--444, 2015.

\bibitem{F63}
Z.~Li, Y.~Yang, J.~Liu, X.~Zhou, and H.~Lu.
\newblock Unsupervised feature selection using nonnegative spectral analysis.
\newblock {\em in Proc. AAAI 2012}, pages 1026--1032, 2012.

\bibitem{F16}
J.~Lines and A.~Bagnall.
\newblock Time series classification with ensembles of elastic distance
  measures.
\newblock {\em Data Min. Knowl. Disc.}, 29:565--592, 2015.

\bibitem{F17}
J.~Lines, S.~Taylor, and A.~Bagnall.
\newblock Time series classification with hive-cote: The hierarchical vote
  collective of transformation-based ensembles.
\newblock {\em ACM Trans. Knowl. Discov. D.}, 21(52):1--35, 2018.

\bibitem{F32}
F.~{Liu}, X.~{Zhou}, J.~{Cao}, Z.~{Wang}, T.~{Wang}, H.~{Wang}, and Y.~{Zhang}.
\newblock Anomaly detection in quasi-periodic time series based on automatic
  data segmentation and attentional lstm-cnn.
\newblock {\em IEEE Trans. Knowl. Data En.,}, pages 1--14, 2020.

\bibitem{F15}
Q.~Ma, J.~Zheng, S.~Li, and G.~W. Cottrell.
\newblock Learning representations for time series clustering.
\newblock {\em in Proc. NeurIPS 2019}, pages 1--11, 2019.

\bibitem{F71}
Q.~Ma, W.~Zhuang, S.~Li, D.~Huang, and G.~W. Cottrell.
\newblock Adversarial dynamic shapelet networks.
\newblock {\em in Proc. AAAI 2020}, pages 5069--5076, 2020.

\bibitem{F19}
L.~Maasten.
\newblock Learning discriminative fisher kernels.
\newblock {\em in Proc. ICML 2011}, pages 217--224, 2011.

\bibitem{F48}
N.~S. Madiraju, S.~M. Sadat, D.~Fisher, and H.~Karimabadi.
\newblock Deep temporal clustering: fully unsupervised learning of time-domain
  features.
\newblock {\em arXiv preprint arXiv: 1802.01059}, 2018.

\bibitem{F3}
R.H. Mousavi, M.~Khansari, and R.~Rahmani.
\newblock A fully scalable big data framework for botnet detection based on
  network traffic analysis.
\newblock {\em Inform. Sciences}, 512:629--640, 2020.

\bibitem{F68}
J.~Paparrizos and L.~Gravano.
\newblock K-shape: efficient and accurate clustering of time series.
\newblock {\em in Proc. ACM SIGMOD 2015}, pages 1855--1870, 2015.

\bibitem{F20}
W.~{Pei}, H.~{Dibeklio{\u g}lu}, D.~M.~J. {Tax}, and L.~{van der Maaten}.
\newblock Multivariate time-series classification using the hidden-unit
  logistic model.
\newblock {\em IEEE Trans. Neur. Net. Lear.}, 29(4):920--931, 2018.

\bibitem{F67}
F.~Petitjean, A.~Ketterlin, and P.~Gancarski.
\newblock A global averaging method for dynamic time warping, with the
  applications to clustering.
\newblock {\em Pattern Recogn.}, 44(3):678--693, 2011.

\bibitem{F10}
M.~Pontiki, D.~Galanis, J.~Pavlopoulos, H.~Papageorgiou, I.~Androutsopoulos,
  and S.~Manandhar.
\newblock Semeval-2014 task 4: Aspect based sentiment analysis.
\newblock {\em in Proc. 8th SemEval 2014}, 2014.
\newblock URL http://alt.qcri.org/semeval2014/task4/.

\bibitem{F53}
A.~R. Puiz, M.~Flynn, and A.~Bagnall.
\newblock Benchmarking multivariate time series classification algorithms.
\newblock {\em arXiv preprint arXiv:2007.13156}, 2020.

\bibitem{F8}
R.~Pérez-Chacón, G.~Asencio-Cortés, F.~Martínez-Álvarez, and A.~Troncoso.
\newblock Big data time series forecasting based on pattern sequence similarity
  and its application to the electricity demand.
\newblock {\em Inform. Sciences}, 540:160–174, 2020.

\bibitem{F64}
M.~Qian and C.~Zhai.
\newblock Robust unsupervised feature selection.
\newblock {\em in Proc. IJCAI 2013}, pages 1621--1627, 2013.

\bibitem{F21}
A.~{Quattoni}, S.~{Wang}, L.~{Morency}, M.~{Collins}, and T.~{Darrell}.
\newblock Hidden conditional random fields.
\newblock {\em IEEE Trans. Pattern Anal.}, 29(10):1848--1852, 2007.

\bibitem{F34}
P.~Rajpurkar, A.~Y. Hannun, M.~Haghpanahi, C.~Bourn, and A.~Y. Ng.
\newblock Cardiologist-level arrhythmia detection with convolutional neural
  networks.
\newblock {\em arXiv:1707.01836}, 2017.

\bibitem{F50}
W.~M. Rand.
\newblock Objective criteria for the evaluation of clustering methods.
\newblock {\em J. Am. Stat. Assoc.}, 166(336):845--850, 1971.

\bibitem{F46}
J.~Redmon and A.~Farhadi.
\newblock Yolov3: An incremental improvement.
\newblock {\em arXiv preprint arXiv: 1804.02767}, 2018.

\bibitem{F42}
F.~Rodrigues, I.~Markou, and F.~C. Pereira.
\newblock Combining time-series and textual data for taxi demand prediction in
  event areas: a deep learning approach.
\newblock {\em Inform. Fusion}, 49:120--129, 2019.

\bibitem{F75}
S.~Ruder.
\newblock An overview of gradient descent optimization algorithms.
\newblock {\em arXiv preprint arXiv: 1609.04747v2}, 2017.

\bibitem{F56}
P.~Schäfer and U.~Leser.
\newblock Multivariate time series classification with weasel+muse.
\newblock {\em arXiv preprint arXiv:1711.11343}, 2017.

\bibitem{F35}
J.~Serrà, S.~Pascual, and A.~Karatzoglou.
\newblock Towards a universal neural network encoder for time series.
\newblock {\em in Proc. CCIA 2018}, pages 120--129, 2018.

\bibitem{F65}
L.~Shi, L.~Du, and Y.~Shen.
\newblock Robust spectral learning for unsupervised feature selection.
\newblock {\em in Proc. IEEE ICMD 2014}, pages 977--982, 2014.

\bibitem{F57}
M.~Shokoohi-Yekta, B.~Hu, H.~Jin, J.~Wang, and E.~Keogh.
\newblock Generalizing dtw to the multi-dimensional case requires an adaptive
  approach.
\newblock {\em Data Min. Knowl. Disc.}, 31:1--31, 2017.

\bibitem{F37}
K.~Shuang, Z.~Zhang, J.~Loo, and S.~Su.
\newblock Convolution-deconvolution word embedding: an end-to-end
  multi-prototype fusion embedding method for natural language processing.
\newblock {\em Inform. Fusion}, 53:112--122, 2020.

\bibitem{F45}
C.~{Siridhipakul} and P.~{Vateekul}.
\newblock Multi-step power consumption forecasting in thailand using dual-stage
  attentional lstm.
\newblock {\em in Proc. IEEE ICITEE 2019}, pages 1--6, 2019.

\bibitem{F25}
W.~Tang, G.~Long, L.~Liu, T.~Zhou, J.~Jiang, and M.~Blumenstein.
\newblock Rethinking 1d-cnn for time series classification: a stronger
  baseline.
\newblock {\em arXiv preprint arXiv: 2002.10061}, 2020.

\bibitem{F60}
K.~S. Tuncel and M.~G. Baydogan.
\newblock Autoregressive forests for multivariate time series modeling.
\newblock {\em Pattern Recogn.}, 73:202--215, 2018.

\bibitem{F28}
A.~Vaswani, N.~Shazeer, N.~Parmar, J.~Uszkoreit, L.~Jones, A.~N. Gomez,
  {\L}.~Kaiser, and I.~Polosukhin.
\newblock Attention is all you need.
\newblock {\em in Proc. NeurIPS 2017}, pages 5998--6008, 2017.

\bibitem{F11}
D.~Wang, R.~Zhang, M.~Liao, M.~Yang, B.~Shi, X.~Liu, Y.~Zhou, D.~Karatzas,
  S.~Lu, C.~V. Jawahar, and X.~Bai.
\newblock Icdar 2019 robust reading challenge on reading chinese text on
  signboard.
\newblock {\em in Proc. ICDAR 2019 robust reading competitions}, 2019.
\newblock URL https://rrc.cvc.uab.es/?com=introduction.

\bibitem{F39}
Q.~{Wang}, C.~{Yuan}, and Z.~{Lin}.
\newblock Learning attentional recurrent neural network for visual tracking.
\newblock {\em in Proc. ICME 2017}, pages 1237--1242, 2017.

\bibitem{F2}
W.~Wang, Y.~Shang, Y.~He, Y.~Li, and J.~Liu.
\newblock Botmark: Automated botnet detection with hybrid analysis of
  flow-based and graph-based traffic behaviors.
\newblock {\em Inform. Sciences}, 511:284--296, 2020.

\bibitem{F33}
Z.~Wang, W.~Yan, and T.~Oates.
\newblock Time series classification from scratch with deep neural networks: A
  strong baseline.
\newblock {\em in Proc. IEEE IJCNN 2017}, pages 1578--1585, 2017.

\bibitem{F59}
M.~Wistuba, J.~Grabocka, and L.~Schmidt-Thieme.
\newblock Ultra-fast shapelets for time series classification.
\newblock {\em arXiv preprint arXiv:1503.05018}, 2015.

\bibitem{F70}
J.~Xie, R.~Girshick, and A.~Farhadi.
\newblock Unsupervised deep embedding for clustering analysis.
\newblock {\em in Proc. ICML 2016}, pages 478--487, 2016.

\bibitem{F66}
J.~Yang and J.~Leskovec.
\newblock Patterns of temporal variation in online media.
\newblock {\em in Proc. ACM WSDM 2011}, pages 177--186, 2011.

\bibitem{F62}
Y.~Yang, H.~Shen, Z.~Ma, Z.~Huang, and X.~Zhou.
\newblock L2, 1-norm regularized discriminative feature selection for
  unsupervised.
\newblock {\em in Proc. IJCAI 2011}, pages 1589--1594, 2011.

\bibitem{F38}
Q.~Yao, R.~Wang, X.~Fan, J.~Liu, and Y.~Li.
\newblock Multi-class arrhythmia detection from l2lead varied-length ecg using
  attention-based time-incremental convolutional neural network.
\newblock {\em Inform. Fusion}, 53:174--182, 2020.

\bibitem{F69}
J.~Zakaris, A.~Mueen, and E.~Keogh.
\newblock Clustering time series using unsupervised shapelets.
\newblock {\em in Proc. IEEE ICMD 2012}, pages 785--794, 2012.

\bibitem{F5}
H.~Zhang, Z.~Xiao, J.~Wang, F.~Li, and E.~Szczerbicki.
\newblock A novel iot-perceptive human activity recognition {(HAR)} approach
  using multi-head convolutional attention.
\newblock {\em {IEEE} Internet Things J.}, 7(2):1072--1080, 2020.

\bibitem{F1}
S.~Zhang, Y.~Chen, W.~Zhang, and R.~Feng.
\newblock A novel ensemble deep learning model with dynamic error correction
  and multi-objective ensemble pruning for time series forecasting.
\newblock {\em Inform. Sciences}, 544:427--445, 2021.

\bibitem{F41}
W.~Zhang, J.~Yu, H.~Hu, and Z.~Qin.
\newblock Multimodal feature fusion by relational reasoning and attention for
  visual question answering.
\newblock {\em Inform. Fusion}, 55:116--126, 2020.

\bibitem{F58}
X.~Zhang, Y.~Gao, J.~Lin, and C.-T. Lu.
\newblock Tapnet: Multivariate time series classification with attentional
  prototypical network.
\newblock {\em in Proc. AAAI 2020}, pages 6845--6852, 2020.

\bibitem{F43}
Y.~Zhu, C.~Zhao, H.~Guo, J.~Wang, X.~Zhao, and H.~Lu.
\newblock Attention couplenet: fully convolutional attention coupling network
  for object detection.
\newblock {\em IEEE Trans. Image Process.}, 28(1):1170--1175, 2018.

\end{thebibliography}

\end{document}